\documentclass[journal]{IEEEtran}
\usepackage{amsmath,amsfonts}
\usepackage{algorithmic}
\usepackage{array}
\usepackage[caption=false,font=normalsize,labelfont=sf,textfont=sf]{subfig}
\newtheorem{remark}{Remark}
\usepackage{textcomp}
\usepackage{stfloats}
\usepackage{url}
\usepackage{verbatim}
\usepackage{graphicx}
\usepackage{hyperref}  
\usepackage{booktabs}
\usepackage{multirow}
\usepackage{colortbl} 
\usepackage[numbers,sort&compress]{natbib}
\usepackage[table,xcdraw]{xcolor}
\def\BibTeX{{\rm B\kern-.05em{\sc i\kern-.025em b}\kern-.08em
    T\kern-.1667em\lower.7ex\hbox{E}\kern-.125emX}}
\usepackage{balance}
\begin{document}
\title{\huge Without Paired Labeled Data: End-to-End Self-Supervised Learning for Drone-view Geo-Localization} 
\author{
\thanks{}}
\author{Zhongwei Chen, \textit{Student Member, IEEE}, Zhao-Xu Yang, \textit{Member, IEEE}, Hai-Jun Rong,  \textit{Senior Member, IEEE}, Guoqi Li, \textit{Member, IEEE}\\
\thanks{This paper is submitted for review on \today. This work was supported in part by the Key Research and Development Program of Shaanxi, PR China (No. 2023-YGBY-235), the National Natural Science Foundation of China (No. 61976172 and No. 12002254), Major Scientific and Technological Innovation Project of Xianyang, PR China (No. L2023-ZDKJ-JSGG-GY-018). (Corresponding author: Zhao-Xu Yang and Hai-Jun Rong)}

\thanks{Zhongwei Chen, Zhao-Xu Yang and Hai-Jun Rong  are with the State Key Laboratory for Strength and Vibration of Mechanical Structures, Shaanxi Key Laboratory of Environment and Control for Flight Vehicle, School of Aerospace Engineering, Xi’an Jiaotong University, Xi’an 710049, PR China (e-mail:ISChenawei@stu.xjtu.edu.cn; yangzhx@xjtu.edu.cn; hjrong@mail.xjtu.edu.cn).

Guoqi Li is with the Institute of Automation, Chinese Academy of Sciences, Beijing 100190, China, also with the School of Artificial Intelligence,
University of Chinese Academy of Sciences, Beijing 101408, China, and
also with the Peng Cheng Laboratory, Shenzhen 518000, China (e-mail:
guoqi.li@ia.ac.cn).}}

\maketitle

\begin{abstract}
Drone-view Geo-Localization (DVGL) aims to achieve accurate localization of drones by retrieving the most relevant GPS-tagged satellite images. However, most existing methods heavily rely on strictly pre-paired drone-satellite images for supervised learning. When the target region shifts, new paired samples are typically required to adapt to the distribution changes. The high cost of annotation and the limited transferability of these methods significantly hinder the practical deployment of DVGL in open-world scenarios. To address these limitations, we propose a novel end-to-end self-supervised learning method with a shallow backbone network, called the dynamic memory-driven and neighborhood information learning (DMNIL) method. It employs a clustering algorithm to generate pseudo-labels and adopts a dual-path contrastive learning framework to learn discriminative intra-view representations. Furthermore, DMNIL incorporates two core modules, including the dynamic hierarchical memory learning (DHML) module and the information consistency evolution learning (ICEL) module. The DHML module combines short-term and long-term memory to enhance intra-view feature consistency and discriminability. Meanwhile, the ICEL module utilizes a neighborhood-driven dynamic constraint mechanism to systematically capture implicit cross-view semantic correlations, consequently improving cross-view feature alignment. To further stabilize and strengthen the self-supervised training process, a pseudo-label enhancement strategy is introduced to enhance the quality of pseudo supervision. Extensive experiments on three public benchmark datasets demonstrate that the proposed method consistently outperforms existing self-supervised methods and even surpasses several state-of-the-art supervised methods. {Our code is available at {\color{red}{\href{https://github.com/ISChenawei/DMNIL}{https://github.com/ISChenawei/DMNIL}}}}.
\end{abstract}


\section{Introduction}\label{introduction}
\begin{figure*}[t]
  \includegraphics[width=7.1in]{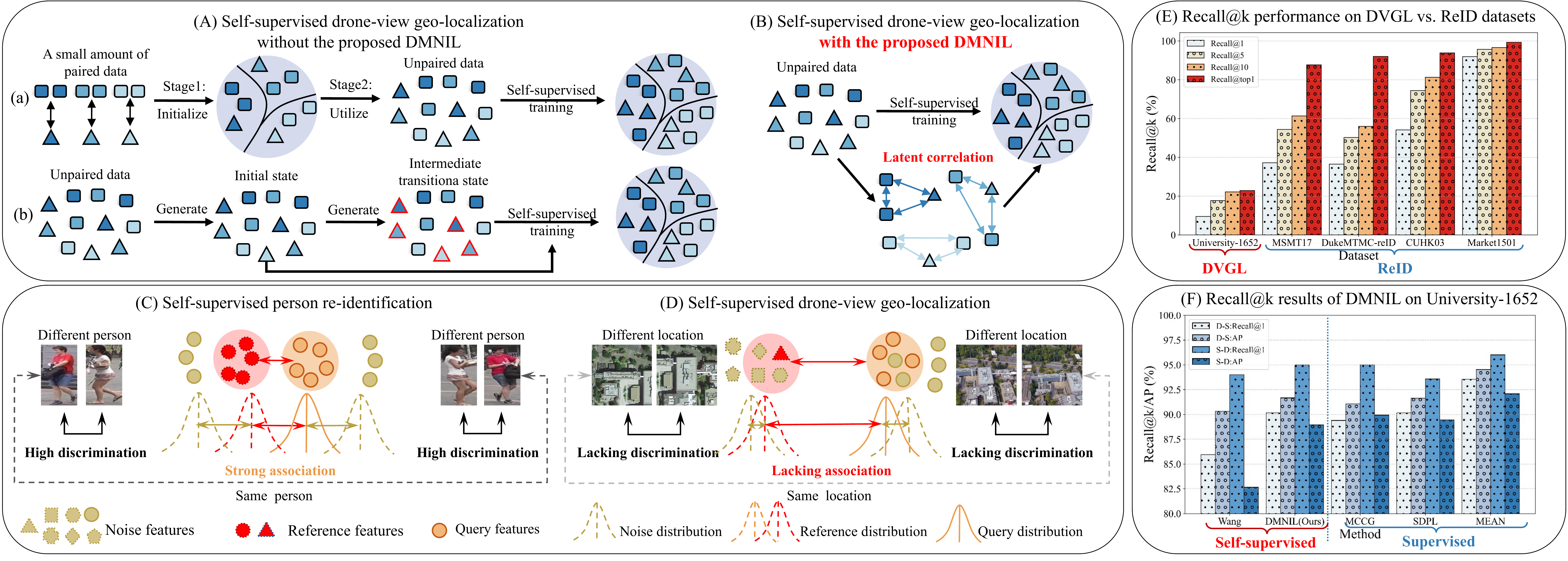}
    \caption{\textbf{Motivation, Challenges, and Evaluation}. (A-B): Motivation of the proposed DMNIL, which aims to explore the latent relationships between cross-view features, without relying on any paired data and multi-stage training and seeking intermediate transitional states to bridge the cross-view gap. (C-D): As representative cross-view retrieval tasks, ReID and DVGL share the common objective of establishing correspondences between query and gallery images based on discriminative feature representations. The notable progress achieved by ReID, especially in self-supervised learning scenarios, offers a valuable methodological foundation and insightful paradigms for advancing DVGL research. However, DVGL presents greater challenges for effective feature learning and discrimination compared to ReID. (C): ReID tasks are characterized by relatively small intra-class variations and distinct inter-class differences, which facilitates effective feature discrimination. (D): In contrast, DVGL tasks suffer from large intra-class variations and small inter-class differences, leading to cross-view gaps and intra-view ambiguities that hinder feature discrimination. (E): The Recall@k performance of the solely used backbone ConvNeXt-Tiny \cite{liu2022convnet} that is pre-trained on ImageNet-22k\cite{deng2009imagenet} is presented on two benchmarks, where the evaluation is conducted on the DVGL dataset University-1652\cite{zheng2020university} and the four person ReID datasets including  MSMT17\cite{wei2018person}, DukeMTMC-reID\cite{zheng2017discriminatively}, CUHK03\cite{sun2014deep}, and Market1501\cite{zheng2015scalable}. (F): Performance of DMNIL is compared with state-of-the-art self-supervised and supervised methods, which outperforms existing state-of-the-art self-supervised methods and even surpasses several supervised methods.}
  \label{fig1}
  \end{figure*}

\IEEEPARstart{D}{rone-View} Geo-Localization (DVGL) is intended to achieve high-precision localization of drones in GPS-denied scenarios by retrieving GPS-tagged satellite images that correspond to drone aerial images \cite{shen2023mccg}. This capability is critical for applications such as autonomous navigation, disaster rescue, and emergency response. In recent years, existing DVGL methods have mainly relied on supervised learning using strictly paired drone-satellite images \cite{chen2024multi, xia2024enhancing}. Despite the promising performance demonstrated by these supervised methods, several intrinsic limitations are exposed. Firstly, the acquisition of instance-level paired drone-satellite images incurs additional training expenses \cite{li2024learning}.{\color{black}{ Despite the promising performance demonstrated by these supervised methods, several intrinsic limitations are exposed. Firstly, the acquisition of instance-level paired drone-satellite images incurs additional training expenses. Automatically pairing drone-satellite images using GPS is infeasible due to unreliable GPS signals and severe viewpoint discrepancies between drone and satellite imagery, so existing datasets \cite{zheng2020university,zhu2023sues,wang2024multiple} rely heavily on manual screening and multi-stage human verification to ensure accurate cross-view.}} Secondly, large amounts of unpaired cross-view image resources remain under-exploited. Lastly, when these methods are applied to new scenarios, the need arises for re-collection and re-pairing of drone-satellite images for training purposes to accommodate domain shifts. These limitations hinder the practical deployment of DVGL. Although researchers have begun to explore self-supervised DVGL methods to address these challenges, existing methods remain limited. The existing method \cite{chen2025limited} attempted to discover cross-view relationships from unpaired data through multi-stage training initialized with a small amount of paired data, as shown in Fig. \ref{fig1}(A(a)). As shown in Fig. \ref{fig1}(A(b)), concurrent work \cite{wang2025coarse} attempted to use intermediate states to bridge the cross-view gap. In contrast, our method aims to directly and effectively uncover latent cross-view correlations in an end-to-end manner without intermediate states, as illustrated in Fig. \ref{fig1}(B).

In recent years, self-supervised person re-identification (ReID) methods \cite{yin2023real,10680455} have achieved remarkable progress to offer crucial theoretical foundations for DVGL. Typically, these methods employ clustering algorithms \cite{ester1996density, macqueen1967some} on features extracted by a backbone network to generate pseudo-labels, which serve as the initial supervision signals for self-supervised learning. The clustering-based pseudo-label generation solution has been widely adopted and has proven highly effective in self-supervised ReID. Inspired by this, we adopt a similar solution in our method to obtain pseudo-labels. However, as illustrated in Fig. \ref{fig1}(C) and Fig. \ref{fig1}(D), compared with conventional ReID tasks, DVGL presents more severe challenges, such as large intra-view ambiguities and cross-view discrepancies. As shown in Fig. \ref{fig1}(E), these issues hinder the backbone network from extracting stable and discriminative features at the early stage, which directly undermines the quality of pseudo-labels generated during the initial clustering phase. To address these, it would be essential to design effective strategies to guide drone features and satellite features to achieve effective intra-class cohesion and inter-class separation within their respective views, which can form distinct clusters. Simultaneously, cross-view consistency learning process would be deliberately optimized to preserve cross-view feature correlations. In the absence of such guidance, the model lacks the necessary ability to learn meaningful representations from low-quality pseudo-labels. This deficiency can trigger a degenerative feedback loop, wherein inaccurate pseudo-labels undermine the quality of feature representations to further compromise pseudo-label quality in subsequent iterations, ultimately leading to the collapse of self-supervised training. 

In this paper, we propose a novel end-to-end self-supervised learning method for DVGL, namely the dynamic memory-driven and neighborhood information learning (DMNIL) method. A lightweight and shallow ConvNeXt-Tiny backbone network \cite{liu2022convnet} is employed to extract features, which are subsequently clustered to generate pseudo-labels. Combined with a dual-path contrastive learning strategy \cite{dai2022cluster,adca}, the backbone network coupled with a downstream clustering process forms the baseline network of DMNIL for learning intra-view feature representations. Building upon this baseline, we design the dynamic hierarchical memory learning (DHML) module and the information consistency evolution learning (ICEL) module as two key components of DMNIL. The DHML module captures the relationship between short-term and long-term memory to enhance the discriminability of intra-view feature representations. The ICEL module constructs a neighborhood-driven dynamic constraint mechanism by integrating neighborhood consistency constraints with mutual information optimization to facilitate cross-view feature alignment and discriminative representation learning. DHML enhances intra-view discrimination and ICEL improves cross-view consistency. Through joint optimization of intra-view and cross-view features, they collectively strengthen the feature discriminability and consistency. Additionally, a pseudo-label enhancement (PLE) strategy is employed to improve the reliability of pseudo-labels and guide more stable self-supervised feature learning. DMNIL surpasses the state-of-the-art self-supervised approaches and achieves performance comparable to supervised methods, as illustrated in Fig. \ref{fig1}(F). The main contributions can be summarized as follows.

\begin{itemize}
  \item We propose an effective dynamic memory-driven and neighborhood information learning method for self-supervised DVGL. Built upon a shallow backbone and a dual-path contrastive learning strategy, our method integrates dynamic hierarchical memory learning and information consistency evolution learning to enhance feature consistency and discriminability for both intra-view and cross-view representation learning. Additionally, a PLE strategy is employed to guide more stable self-supervised learning.
  \item To address the intra-view ambiguity caused by spatial continuity across geographic viewpoints, we propose a dynamic memory hierarchical learning (DHML) module that captures the dependencies between different memory levels to adaptively learn the dynamics of intra-class and inter-class features within the intra-view.
  \item To address cross-view discrepancies and inconsistencies, we propose a novel information consistency evolution learning (ICEL) module which combines neighborhood consistency and mutual information optimization to learn consistent and discriminative cross-view representations, bridging the discriminability in intra-view learned by the DHML module to form a synergistic learning system.
  \item Extensive experiments demonstrate that DMNIL, which employs a lightweight backbone, outperforms existing self-supervised methods on three benchmarks and even surpasses several state-of-the-art supervised methods.
\end{itemize}
The remainder of this paper is organized as follows. Section \ref{related works} systematically reviews previous research. In Section \ref{method}, the proposed self-supervised learning method is presented in detail. The experimental results are reported and analyzed in Section \ref{results}. Finally, conclusions are outlined in Section \ref{conclusions}.

\section{Related Works}\label{related works}
This section establishes a systematic and comprehensive theoretical framework to support the development of the self-supervised learning method proposed for DVGL. DVGL is typically regarded as a sub-task of cross-view geo-localization (CVGL). To provide a more comprehensive understanding of the field, we first review the overall progress in CVGL, including both supervised and self-supervised methods. Then, we summarize the main methods in self-supervised ReID that have provided analogous problem-solving paradigms.
\subsection{Supervised Cross-View Geo-localization}
CVGL aims to achieve high-precision localization of query images by retrieving GPS-tagged satellite images. Early research focused mainly on the CVGL in ground-view \cite{10601183} with several benchmark datasets, such as CVUSA \cite{zhai2017predicting}, CVACT \cite{2019Lending}, and VIGOR \cite{zhu2021vigor}. To address the challenge of ground-view CVGL, researchers have proposed various methods. CVM-Net \cite{Hu_2018_CVPR} employed NetVLAD \cite{arandjelovic2016netvlad} to encode images as global descriptors to reduce visual discrepancies caused by viewpoint variations. TransGeo \cite{zhu2022transgeo} utilized attention-based zooming and sharpness-aware optimization to enhance detail learning and improve model generalization. GeoDTR \cite{zhang2023cross} leveraged a Transformer-based \cite{liu2022swin} feature extractor to disentangle geometric features, which effectively address cross-view shifts. Sample4Geo \cite{deuser2023sample4geo} introduced a hard negative mining strategy, focusing on challenging negative samples to further improve discriminative capability.

With the continuous development and application of drone technology, DVGL has gradually become an important research direction and has achieved significant progress \cite{wu2024camp,lv2024direction,du2024ccr}. Several benchmark datasets for DVGL have been proposed, such as University-1652 \cite{zheng2020university}, SUES-200 \cite{zhu2023sues}, and DenseUAV \cite{DenseUAV}, which have driven further research in this field. In this scenario, researchers have proposed a variety of methods. LPN \cite{wang2021each} extracted fine-grained features through local pattern partitioning to improve the ability to capture local details. IFS \cite{ge2024multibranch} utilized a multi-branch strategy to effectively fuse global and local features. CAMP \cite{wu2024camp} improved the accuracy of cross-view matching by contrastive attribute mining and position-aware partitioning, while DAC \cite{xia2024enhancing} enhanced feature consistency and matching accuracy by introducing domain alignment and scene consistency constraints. MEAN \cite{chen2024multi} adopted progressive embedding diversification, global-to-local associations, and cross-domain enhanced alignment to further improve the representation and alignment of features.

Nevertheless, the remarkable performances exhibited by these DVGL methods require extensive pre-processed drone-satellite image pairs. In this work, we concentrate on self-supervised learning method for DVGL, particularly addressing scenarios where labeled pairings are unavailable, which is crucial for real-world applications.
\subsection{Self-Supervised Cross-View Geo-localization}
Due to the dependence of CVGL on extensive pre-processed image pairs, which significantly increases costs and limits applications for the real world, researchers have begun to explore self-supervised learning methods for CVGL. The work in \cite{Li_2024_CVPR} proposed a self-supervised learning method that used correspondence-free projection to transform ground panorama images into bird's eye view images. It also applied CycleGAN \cite{Zhu_2017_ICCV} to generate fake satellite images, coupled with an additional SAFA module \cite{shi2019spatial} to reduce discrepancies. However, this explicit alignment method has inherent limitations in capturing the deep semantic relationships of cross-view features and may introduce noise. Furthermore, the non-end-to-end training method employed increases training time and computational costs with low practical applicability. Additionally, existing work \cite{li2024learning} utilized frozen foundation models (FMs) \cite{oquab2023dinov2} to extract features, and a self-supervised adapter was introduced to mitigate visual discrepancies between different points of view. However, this method heavily relied on the ability of FMs for feature extraction. When FMs fail to effectively handle viewpoint discrepancies, the performance of the adapter is also affected. To reduce reliance on the feature extractor, CDIKTNet \cite{chen2025limited} trained with a small amount of paired drone-satellite images to optimize the initial feature distribution and then performed clustering to generate high-quality pseudo-labels. However, this method inherently depends on the feature extractor, as it still requires limited paired drone-satellite images for effective learning. 

The above methods have made meaningful exploratory attempts. However, these methods still face challenges, such as the additional computation time and overhead introduced by non-end-to-end frameworks, as well as dependence on the capability of feature extractor. In contrast, DMNIL is fully end-to-end, eliminating the need for additional training processes and completely removing the reliance on feature extractor initialization.

\begin{figure*}[t]
  \centering
  \includegraphics[width=7in]{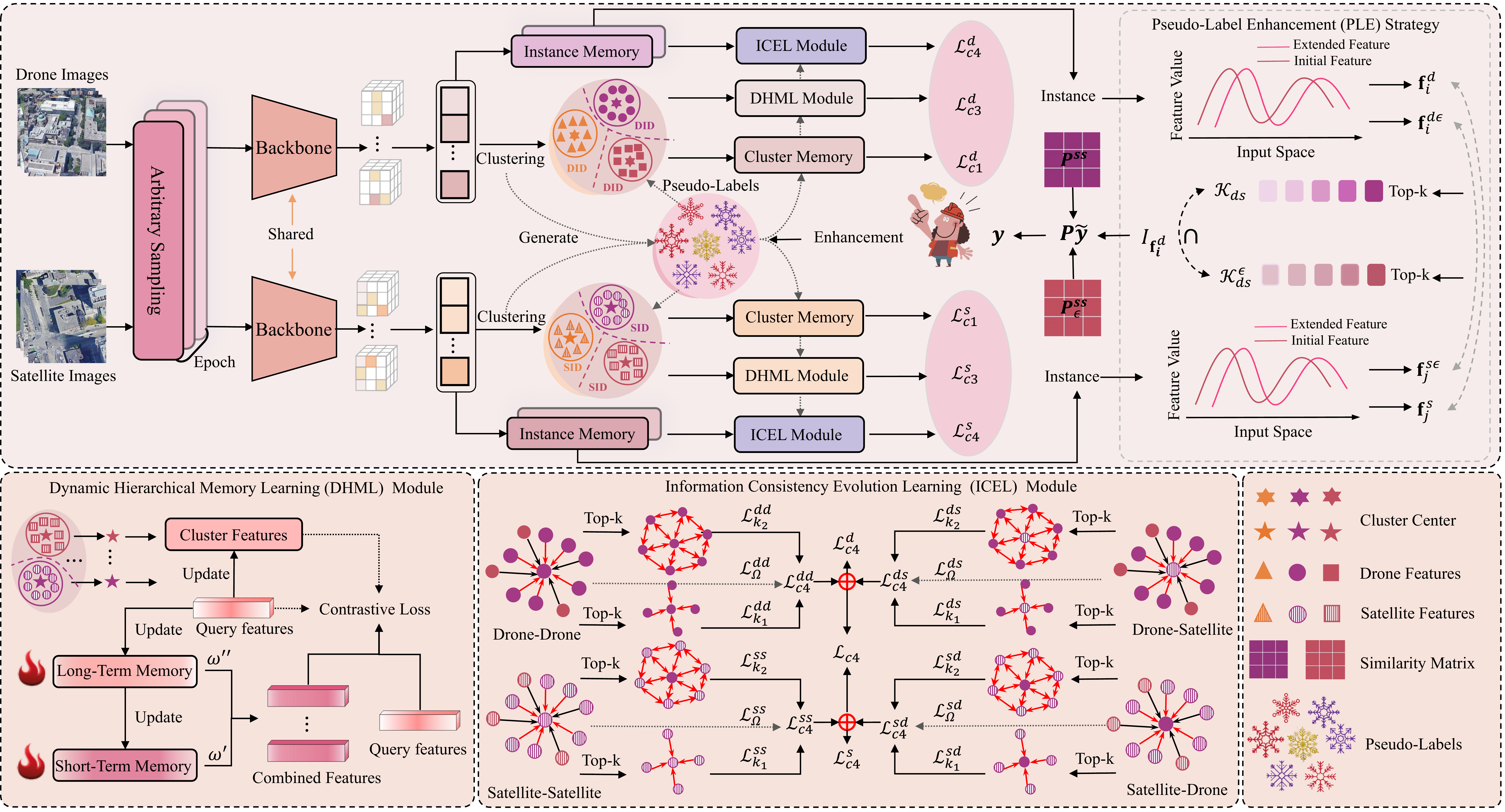}
  \caption{\textbf{Pipeline Overview.} The DMNIL consists of a lightweight backbone, a dual-path contrastive learning strategy, a dynamic hierarchical memory learning module, and an information consistency evolution module. Specifically, the dual-path contrastive learning is designed to learn discriminative and consistent intra-view feature representations. The dynamic hierarchical memory module further captures intra-view feature variations under different viewpoints and scales, thereby enhancing the robustness and discriminability of the learned representations. The information consistency evolution module focuses on modeling cross-view feature consistency through a neighborhood-driven learning strategy, and further improves the training process by integrating a pseudo-label enhancement strategy.}
  \label{fig2}
  \end{figure*} 
\subsection{Methodological Parallels from Self-Supervised ReID}
ReID \cite{9457243,10310270} aims to match the same person image captured by non-overlapping cameras. Due to its critical role in video surveillance, ReID has attracted extensive research attention and achieved remarkable progress. It has provided a fully end-to-end framework to eliminate the need for additional training processes and completely remove the reliance on feature extractor initialization, which offers strong inspiration for our consequent work in the DVGL task. However, these achievements are facilitated by extensive human-labeled data. To address this limitation, self-supervised learning methods \cite{yin2023real,9840394} have been proposed. Among them, clustering algorithms are widely employed to generate pseudo-labels, which serve as supervision signals for model training. In addition, benefitting from the introduction of instance-level contrastive learning \cite{Wu_2018_CVPR} with memory banks and momentum update strategies \cite{He_2020_CVPR}, many methods \cite{chen2021ice,wang2022optimal} consider each unlabeled sample as an individual class to learn discriminative instance-level feature representations. However, these instance-level features were largely modified during training, which resulted in unstable representations \cite{10882953}. To address this issue, recent studies \cite{Yang_2023_ICCV,cheng2023efficient} have begun exploring cluster-level relationships to discover more stable and robust associations among pedestrian representations. Cluster-Contrast \cite{dai2022cluster} contrasts instance features with these cluster-level representations to address the issue of inconsistent instance discrimination. Nevertheless, in cross-modal ReID scenarios, where images are captured by heterogeneous sensors, modality discrepancies hinder the effectiveness of these methods in learning modality-invariant features. ADCA \cite{adca} introduced a dual-path contrastive framework to bridge modality gaps and enhance cross-modal feature alignment. Building on this, SDCL \cite{yang2024shallow} employed a dual-path Transformer for shallow-to-deep contrastive learning, and refined pseudo-labels via consistency constraints across feature hierarchies to achieve more stable and discriminative cross-modal representations. The above methods focus on alleviating the dependency on human-labeled data in the ReID and provide valuable insights for DVGL. However, DVGL suffers from more severe viewpoint and scale variations. Furthermore, the spatial continuity of geographic environments inevitably induces feature ambiguity, which would severely degrade clustering effectiveness. 

A cluster-based self-supervised framework, which is developed concurrently with our work, is proposed with a similar ReID framework that employs intermediate state transition and local-to-global matching for cross-view alignment to mitigate viewpoint discrepancies. However, this method relies on explicit appearance-level transformations rather than intrinsically modeling cross-view semantic features, and lacks end-to-end optimization and uncertainty-aware pseudo-label refinement, making it sensitive to noise and limiting its adaptability. In this work, we integrate instance-level and cluster-level features in a self-supervised method for DVGL by employing reliable neighbor learning to facilitate implicit cross-view alignment and improve the discriminative power and robustness of the learned representations.

\section{End-to-End Self-Supervised Learning Method}\label{method}
As illustrated in Fig.~\ref{fig2}, the proposed method is composed of two key modules, DHML and ICEL. Additionally, a PLE strategy is employed to optimize the self-supervised learning process. Specifically, we adopt a weight-shared ConvNeXt-Tiny network as the backbone, equipped with the cluster memory and the dual-path contrastive learning strategy to learn intra-view feature representations. To capture the dynamic variations of intra-class features caused by changes in viewpoint and scale, the DHML module establishes a collaborative relationship between short-term and long-term memory. DHML module enables efficient discriminability of intra-view feature representations. Furthermore, the ICEL module leverages instance memory to construct neighborhood-driven dynamic constraints. By integrating neighborhood consistency with mutual information optimization, it guides cross-view features to form consistent and unified representations.

\textit{Problem Formulation}: In DVGL, a set of paired drone-satellite images is denoted as \(\{x^d, x^s\}\), where \(x^d\) represents the drone-view image and \(x^s\) represents the satellite-view image. Unlike the supervised setting, our self-supervised method is developed without using paired drone-satellite images, where the correspondences between drone-view and satellite-view images are entirely unavailable. The objective is to discover reliable cross-view correspondences under such conditions to perform the DVGL task.

\subsection{Dual-Path Contrastive Learning Baseline}\label{baseline}
Cluster-based contrastive learning methods have achieved remarkable success in various fields \cite{dai2022cluster,hu2022feature}. However, in DVGL, the unique challenges of cross-view scenarios, as discussed in Section \ref{introduction}, make it difficult for single-path cluster-based contrastive learning to effectively capture view-specific features. Therefore, we design the baseline network with a dual-path architecture, where drone-view and satellite-view features are extracted using a shared backbone and clustered to assign pseudo-labels. Subsequently, a dual-path contrastive learning strategy \cite{yang2023dual} is employed to enhance intra-view representation learning.

We first introduce the notation for better clarity. Let \( X_d = \{x^d_i\}_{i=1}^N \) denote a set of drone images with \( N \) instances, and \( X_s = \{x^s_j\}_{j=1}^M \) denote a set of satellite images with \( M \) instances. \( F_d = \{\mathbf{f}^d_i\}_{i=1}^N \) and \(F_s = \{\mathbf{f}^s_j\}_{j=1}^M \) represent the feature representations extracted by the shared feature extractor \( \mathcal{F}_{backbone} \) for drone-view and satellite-view, respectively. \( q^d\) and \( q^s\) denote the drone query instance features and satellite query instance features extracted by the feature extractor \( \mathcal{F}_{backbone} \).

\textit{Arbitrary Sampling and Backbone-based Feature Extraction.}  
To simulate real-world scenarios where drone and satellite images are not paired, we independently and arbitrarily sample drone and satellite images at each training epoch to construct unpaired batches. Each image is subsequently processed by a shared backbone network to extract features. The backbone processing procedure can be described as
\begin{equation}
\begin{split}
&\mathbf{f}^d_i = \mathcal{F}_{\text{backbone}}(x^d_i),\quad i =1,\dots, N\\
&\mathbf{f}^s_j = \mathcal{F}_{\text{backbone}}(x^s_j),\quad j =1,\dots, M.
\end{split}
\end{equation}

\textit{Cross-View Cluster Memory Initialization}. At the beginning of each training epoch, the feature memories for both views are initialized into $K$ clusters and $L$ clusters, respectively. Specifically, the cluster centroids of the drone-view \(\{ \phi_1^d, \cdots, \phi_K^d \}\) and the satellite-view \(\{ \phi_1^s, \cdots, \phi_L^s \}\) are stored in the drone memory dictionary \(\mathcal{M}_d\) and the satellite memory dictionary \(\mathcal{M}_s\), respectively, for view-specific memory initialization. This process can be formalized as
\begin{equation}
\mathcal{M}_d\ \leftarrow \{\phi_1^d, \cdots, \phi_K^d \}, \mathcal{M}_s\ \leftarrow \{ \phi_1^s, \cdots, \phi_L^s \}
\label{eq2}
\end{equation}
\begin{equation}
\phi_k^d = \frac{1}{|\mathcal{H}_k^d|} \sum_{\mathbf{f}_i^d \in \mathcal{H}_k^d} \mathbf{f}_i^d, \phi_l^s = \frac{1}{|\mathcal{H}_l^s|} \sum_{\mathbf{f}_j^s \in \mathcal{H}_l^s} \mathbf{f}_j^s
\end{equation}
where \(\mathcal{H}_{k}^{d}\) denotes the $k$-th cluster set in drone-view, \(k=1,\cdots,K\), and \(\mathcal{H}_{l}^{s}\) denotes the $l$-th cluster set in satellite-view, \(l=1,\cdots,L\). \(| \cdot |\) indicates the number of instances per cluster.

\textit{Cross-View Cluster Memory Updating}.
{\color{black} During the training phase, we randomly sample \( P \) different localization points from the training set, with each point comprising a fixed number \( Z \) of instances. This sampling strategy results in a minibatch containing a total of \( P \times Z \) query images. Subsequently, the query features extracted from the minibatch are integrated into the cluster representations via a momentum-based update mechanism. The cluster centroids \(\phi_k^{d}\) and \(\phi_l^{s}\) are updated iteratively as
\begin{equation}
\quad \phi_k^{d} \leftarrow \alpha \phi_k^{d} + (1 - \alpha) q^d,\quad \phi_l^{s} \leftarrow \alpha \phi_l^{s} + (1 - \alpha) q^s 
\label{eq4}
\end{equation}
where \(\alpha\) is the momentum factor and is set to 0.1.}

\textit{Cross-View Cluster Memory Contrastive Loss}.
Given drone and satellite query \( q^d\) and \( q^s\), we compute the contrastive loss for drone-view and satellite-view as
\begin{subequations}
\begin{equation}
\mathcal{L}^{d}_{c1} = - \log \frac{\exp(q^{d} \cdot \phi^{d}_{+} / \tau)}{\sum_{k=0}^{K} \exp(q^{d} \cdot \phi^{d}_k / \tau)} 
\end{equation}
\begin{equation}
\mathcal{L}^{s}_{c1} = - \log \frac{\exp(q^{s} \cdot \phi^{s}_{+} / \tau)}{\sum_{l=0}^{L} \exp(q^{s} \cdot \phi^{s}_l / \tau)}
\end{equation}
\end{subequations}
where \(\tau\) denotes the temperature hyper-parameter. Cluster centroids \(\phi^{d}_{k}\) and \(\phi^{s}_{l}\) are utilized as feature vectors at the cluster level to compute the distances between the query instance and all clusters. Here,  \(\phi^{d}_+\) and \(\phi^{s}_+\) are positive cluster features associated with \(q^{d}\) and \(q^{s}\). The final optimization for the baseline is denoted as
\begin{equation}
\mathcal{L}_{c1} = \mathcal{L}^{d}_{c1} + \mathcal{L}^{s}_{c1}.
\end{equation}

{\color{black} \remark{In our baseline, both views share a common backbone to extract unified feature representations. For each view, features from samples with the high similarity are averaged to create a cluster centroid, which is then used as the representative feature for the cluster. The cluster centroid continuously refined through a momentum-based update strategy as shown in Eq.(\ref{eq4}). Building upon this, we minimize the discrepancy between the predicted similarity distribution of query features and the target cluster centroids to guide the query features closer to the target cluster centroids. Simultaneously, we suppress similarities to non-target cluster centroids to enlarge the margin between distinct semantic clusters.}}

\subsection{Dynamic Hierarchical Memory Learning }
Geographic environments inherently exhibit spatial continuity, where neighboring regions usually present similar visual patterns, while distant regions show significant differences. However, such spatial characteristics often introduce ambiguity in intra-view feature representations. To address this issue, we propose a DHML module that jointly exploits long-term and short-term memory to capture spatial dynamics and enhance the discriminability of intra-view features.

Consistent with the baseline network, we construct dynamic hierarchical memory learning in a dual-path manner, using drone feature extraction as an illustrative example. To build long-term memory, we first construct two novel clustering memory dictionaries according to Eq.(\ref{eq2}), which are used to store short-term \(\mathcal{M}_d^{'}\) and long-term \(\mathcal{M}_d^{''}\) memories, respectively. They are expressed as
\begin{equation}
\mathcal{M}_d^{'}\ \leftarrow \{\varphi_1^{d{'}}, \cdots, \varphi_K^{d{'}} \},\mathcal{M}_d^{''}\ \leftarrow \{\varphi_1^{d{''}}, \cdots, \varphi_K^{d{''}} \}
\end{equation}
where \( \varphi_1^{d{'}}, \dots, \varphi_K^{d{'}} \) represent the short-term memory embeddings that encode the cluster representations of the drone-view, and \( \varphi_1^{d{''}}, \dots, \varphi_K^{d{''}} \) represent the long-term memory embeddings that store historically accumulated stable representations of the drone-view. {\color{black}Since the clustering representations in the short-term and long-term memories adopt a different value of momentum factor from that in Eq.(\ref{eq2}) due to different motivations, their update dynamics and resulting representations are also different. Therefore, we use different symbols to distinguish them.}

{\color{black} Unlike Eq.(\ref{eq4}), we assign different momentum factor to represent the weights between the input features and historical memory during the long-term memory update. Specifically, since the momentum parameter $\xi$ is set to 0.3, the influence of historical memory during the long-term memory updates has been enhanced. The updating procedure can be defined as
\begin{equation}\label{eq8}
\quad \varphi_k^{d{''}} \leftarrow (1 - \xi)\varphi_k^{d{''}} +  \xi  q^{d{''}}
\end{equation}
where \( q^{d{''}} \) represents the drone query instance feature in long-term memory. This obtains a new dictionary of long-term memory.}

In addition, we also introduce an adaptive weight update mechanism to update the short-term memory. In detail, we compute the euclidean distance between the current input and its corresponding long-term memory, then take the average over all samples in the batch. This average distance is then mapped to the range \([0,1]\) via the sigmoid function, which yields an adaptive update coefficient
\begin{equation} \label{eq10}
\beta = \delta\!\left(\frac{1}{B}\sum_{i=1}^{B} \| q^{d{''}} - \varphi_k^{d{''}} \| \right)
\end{equation}
where \(B\) denotes the batch size and \(\delta(\cdot)\) represents the sigmoid function. The short-term memory is subsequently updated as
\begin{equation}\label{eq011}
\quad \varphi_k^{d{'}} \leftarrow \beta \cdot \varphi_k^{d{''}} + (1 - \beta) \cdot \varphi_k^{d{'}}.
\end{equation}

Then, the long-term and short-term memories are fused in a weighted manner to yield the combined feature \(\varphi_k^{db}\) representation,
\begin{equation}\label{eq12}
\varphi_k^{d{b}} = w{''} \cdot \varphi_k^{d{''}} + w{'} \cdot \varphi_k^{d{'}}
\end{equation}
where \(w{''}\) and \(w{'}\) denote the respective fusion weights.
As a result, we obtain a comprehensive memory dictionary that effectively integrates long-term global consistency with localized short-term adaptability.

Given the drone query \( q^d \), we compute the contrastive loss by measuring their similarity to the cluster representations maintained in the memory dictionary by the following equations,
\begin{equation}
\mathcal{L}^{d}_{c2} = - \log \frac{\exp(q^{d} \cdot \varphi^{db}_{+} / \tau)}{\sum_{k=0}^{K} \exp(q^{d} \cdot \varphi^{db}_k / \tau)} 
\end{equation}
where \(\tau\) denotes the temperature hyper-parameter. The combined cluster centroids \(\varphi^{db}_{k}\) are used as feature vectors at the cluster level to calculate the distances between the query instance \(q^{d}\) and all clusters. Here,  \(\varphi^{db}_+\) corresponds to the positive cluster feature associated with \(q^{d}\).
Finally, we introduce a fusion factor \( \lambda_{0} \) to perform a weighted fusion of \(\mathcal{L}^{d}_{c2}\) and \(\mathcal{L}^{d}_{c1}\), aiming to enhance the discriminability of intra-view feature representations and obtain the optimized loss $\mathcal{L}_{c3}^d = \lambda_{0}\mathcal{L}^{d}_{c1}\ + \mathcal{L}^{d}_{c2}$.
Similarly, $\mathcal{L}_{c3}^s$ can be obtained in the same way. Therefore, the dynamic hierarchical memory learning process can be summarized as
\begin{equation}
\mathcal{L}_{c3} = \mathcal{L}^{d}_{c3} + \mathcal{L}^{s}_{c3}.
\end{equation}

{\color{black} \remark{The proposed DHML module differs from traditional single-memory methods \cite{dai2022cluster,adca,yang2024shallow}. DHML is modeled as a dual-cycle process across two temporal scales. The long-term memory accumulates stable global features over a long period and is updated with a small momentum to maintain consistency, while the short-term memory rapidly responds to variations in the current input over a short period to capture local dynamic features. The adaptive coefficient $\beta$ functions as a temporal regulator between the two memories. When the input features deviate significantly from the long-term memory, $\beta$ increases and the short-term memory relies more on the stable long-term representation. When the deviation is small, $\beta$ decreases, allowing the short-term memory to retain more of its recent, immediate information.}}

\subsection{Information Consistency Evolution Learning}
The baseline network and the DHML module primarily focus on intra-view feature learning, lacking the ability to model and associate cross-view features effectively. Therefore, the proposed ICEL module employs a threshold-based and ranking-based neighborhood selection strategy to achieve effective cross-view feature alignment. By promoting the consistency evolution of cross-view feature representations, ICEL enhances both the discriminability and the representation capability of cross-view features.

In the ICEL module, unlike the baseline and DHML module, we employ instance-level features instead of cluster-level representations. To achieve this, we define two instance memory dictionaries, \(\mathcal{I}_{d}\) and \(\mathcal{I}_{s}\), which store the instance features extracted by the shared feature encoder, as described in Section \ref{baseline}. This process can be described as
\begin{equation}
\mathcal{I}_{d} \leftarrow \{ \mathbf{f}_1^d, \dots, \mathbf{f}_N^d \},\mathcal{I}_{s} \leftarrow \{ \mathbf{f}_1^s, \dots, \mathbf{f}_M^s \}
\end{equation}
where \( \mathbf{f}_i^d \), $i =1,\dots N$, and \( \mathbf{f}_j^s \), $j =1,\dots M$, denote the drone and satellite instance features stored in memory dictionaries  \(\mathcal{I}_{d}\) and \(\mathcal{I}_{s}\), respectively.

To facilitate the following description, we take the scenario where the drone-view serves as the query as an illustrative example. The similarity between a given drone query \(q_v^{d}\) that denotes the $v$-th query sample in the drone-view, $v =1,\dots N^b$. $N^b$ denote the batch size of the query $q_d$. Each drone instance \(\mathbf{f}_i^{d}\) and each satellite instance \(\mathbf{f}_j^{{s}}\) in the training set are compared with drone query $q_d$ by computing a similarity score, respectively. This is defined as
\begin{equation}\label{eq16}
\mathcal{S}(q_v^{d},\mathbf{f}_i^{d}) = \frac{q_v^{d} \cdot \mathbf{f}_i^{d}}{\| q_v^{d} \|_2 \| \mathbf{f}_i^{d} \|_2}, \mathcal{S}(q_v^{d},\mathbf{f}_j^{s}) = \frac{q_v^{d} \cdot \mathbf{f}_j^{s}}{\| q_v^{d} \|_2 \| \mathbf{f}_j^{s} \|_2}.
\end{equation}

Then, we adopt a threshold-based neighborhood selection strategy to select reliable neighbors, which can be defined as
\begin{subequations}\label{eq17}
\begin{equation}
\Omega^{dd} = \Big\{ \mathbf{f}_{i}^{d} \,\Big|\, S \big(q_v^{d}, \mathbf{f}_{i}^{d} \big) > \gamma  \max_{i \in \{1, \dots, N\}} S \big(q_v^{d}, \mathbf{f}_{i}^{d}) \Big\}
\end{equation}
\begin{equation}\label{eq18}
\Omega^{ds} = \Big\{ \mathbf{f}_{j}^{s} \,\Big|\, S \big(q_v^{d}, \mathbf{f}_{j}^{s} \big) > \gamma  \max_{j \in \{1, \dots, M\}} S \big(q_v^{d}, \mathbf{f}_{j}^{s} ) \Big\}
\end{equation}
\end{subequations}
where \(\Omega^{dd}\) represents the set of drone feature instances selected as valid intra-view neighbor, and \(\Omega^{ds}\) denotes the set of drone feature instances that exhibit strong cross-view associations with satellite features. \(\gamma\in (0,1)\) is a predefined threshold.

Additionally, we introduce a ranking-based neighborhood selection strategy. Specifically, we independently select the top-\( k_1 \) most relevant samples to construct a strictly filtered neighborhood, ensuring high-confidence associations, while also selecting the top-\( k_2 \) samples define an expanded neighborhood, capturing additional informative relationships. Note that \( k_2 > k_1 \). {\color{black} Here $\mathrm{TopK}_S(q,\mathcal{F},k)$ denotes the ranking-based neighborhood selection strategy, which returns the top-$k$ elements in $\mathcal{F}$ ranked by their similarity to $q$.} The process can be defined as\label{k1andk2}
{\color{black}
\begin{subequations}
\begin{align}
\mathcal{N}_{k_1}^{dd} &= \mathrm{TopK}_S\!\big(q_v^{d}, \{\mathbf{f}_i^{d}\}_{i=1}^N, k_1\big),\\
\mathcal{N}_{k_1}^{ds} &= \mathrm{TopK}_S\!\big(q_v^{d}, \{\mathbf{f}_j^{s}\}_{j=1}^M, k_1\big) ,
\end{align}
\end{subequations}
where \( \mathcal{N}_{k_1}^{dd} \) and \( \mathcal{N}_{k_1}^{ds} \) denote the selected intra-view and cross-view neighborhood sets, respectively. \( \mathcal{N}_{k_2}^{dd} \) and \( \mathcal{N}_{k_2}^{ds} \) can be obtained following the same selection strategy.}

Based on the constructed neighborhoods, the normalized similarity distribution is computed within the larger neighborhood to promote the aggregation of similar samples in both intra-view and cross-view scenarios. This distribution is then optimized by a cross-entropy loss to enhance alignment between query samples and their neighbors. The intra-view \(\mathcal{L}_{\Omega}^{dd}\) and cross-view \(\mathcal{L}_{\Omega}^{ds}\) losses can be formulated as
\begin{subequations}
\begin{equation}
\mathcal{L}_{\Omega}^{dd} \!=\! -\frac{1}{N^b} \sum_{v=1}^{N^b} \sum_{\mathbf{f}_i^{d} \in \Omega^{dd}} \!\log \frac{\exp \left( \mathcal{S}(q_v^{d{}},\mathbf{f}_i^{d}) / \tau \right)}
{\sum_{\mathbf{f}_i^{d} \in {{{\Omega}^{dd}}}} \exp \left( \mathcal{S}(q_v^{d{}},\mathbf{f}_i^{d}) / \tau \right)}
\end{equation}
\begin{equation}
\mathcal{L}_{\Omega}^{ds} \!=\! -\frac{1}{N^b} \sum_{v=1}^{N^b} \sum_{\mathbf{f}_j^{s} \in \Omega^{ds}} \!\log \frac{\exp \left( \mathcal{S}(q_v^{d{}},\mathbf{f}_j^{s}) / \tau \right)}
{\sum_{\mathbf{f}_j^{s} \in {{{\Omega}^{ds}}}} \exp \left( \mathcal{S}(q_v^{d{}},\mathbf{f}_j^{s}) / \tau \right)}
\end{equation}
\end{subequations}
where \( N^{{\Omega}^{dd}} \) and \( N^{{\Omega}^{ds}} \) denote the total number of neighborhood samples within the set \({\Omega}^{dd}\) and \({\Omega}^{ds}\).

Incorporating neighborhoods \( \mathcal{N}_{k_1}^{dd} \), \( \mathcal{N}_{k_2}^{dd} \) and \( \mathcal{N}_{k_1}^{ds} \), \( \mathcal{N}_{k_2}^{ds} \), we introduce consistency loss to enhance alignment of relative similarity distributions within the large-scale neighborhood \( k_2 \) across both intra-view and cross-view representations. The intra-view loss \(\mathcal{L}_{k_2}^{dd}\) and cross-view loss \(\mathcal{L}_{k_2}^{ds}\) under this consistency loss can be formulated as
\begin{subequations}
\begin{equation}
p_{v,i}^{dd} = \frac{\exp(\mathcal{S}(q_v^{d}, \mathbf{f}_{i}^{d}))}{\sum_{\mathbf{f}_{i}^{d}\in \mathcal{N}_{k_2}^{dd}} \exp(S(q_v^{d}, \mathbf{f}_{i}^{d}))}
\end{equation}
\begin{equation}
p_{v,j}^{ds} = \frac{\exp(\mathcal{S}(q_v^{d}, \mathbf{f}_{j}^{s}))}{\sum_{\mathbf{f}_{j}^{s} \in \mathcal{N}_{k_2}^{ds}} \exp(S(q_v^{d}, \mathbf{f}_{j}^{s}))}
\end{equation}
\begin{equation}
\mathcal{L}_{k_2}^{dd} = \frac{1}{N^b} \sum_{v=1}^{N^b} 
\sum_{i \in \mathcal{N}_{k_2}^{dd}} 
p_{v,i}^{dd}
\log \left( k_2 \cdot p_{v,i}^{dd} \right)
\end{equation}
\begin{equation}
\mathcal{L}_{k_2}^{ds} = \frac{1}{N^b} \sum_{v=1}^{N^b} 
\sum_{j \in \mathcal{N}_{k_2}^{ds}} 
p_{v,j}^{ds}
\log \left( k_2 \cdot p_{v,j}^{ds} \right)
\end{equation}
\end{subequations}
where \( p_{v,i}^{dd}\) and  \( p_{v,j}^{ds}\) are the softmax-normalized similarity distribution.

Within a more restrictive and smaller set of \( k_1 \) neighbors, a mutual information constraint is proposed to learn intra-view and cross-view features representations. This complements the consistency constraint described above and achieves a balance between global structural alignment and local discriminative information. Within the selected \(k_1\) neighbors, we compute the similarity between the query \(q_v^{d}\) and each neighbor \(\mathbf{f}_{i}^{d}\) and \(\mathbf{f}_{j}^{s}\) using the function \(\mathcal{S}(\cdot, \cdot)\), and apply softmax to obtain the posterior probability for each neighbor. It can be described as
\begin{subequations}
\begin{equation}
p^{dd}(i|v) = \frac{\exp\big(\mathcal{S}(q_v^{d}, \mathbf{f}_{i}^{d})\big)}{\sum_{\mathbf{f}_{i}^{d}\in \mathcal{N}_{k_1}^{dd}}\exp\big(\mathcal{S}(q_v^{d}, \mathbf{f}_{i}^{d})\big)}
\end{equation}
\begin{equation}
p^{ds}(j|v) = \frac{\exp\big(\mathcal{S}(q_v^{d}, \mathbf{f}_{j}^{s})\big)}{\sum_{\mathbf{f}_{j}^{s}\in \mathcal{N}_{k_1}^{ds}}\exp\big(\mathcal{S}(q_v^{d}, \mathbf{f}_{j}^{s})\big)}.
\end{equation}
\end{subequations}

Therefore, the intra-view and cross-view mutual information losses \(\mathcal{L}_{k_1}^{dd}\) and \(\mathcal{L}_{k_1}^{ds}\) can be derived and are defined as
\begin{subequations}
\begin{equation}
\mathcal{L}_{k_1}^{dd} = - \frac{1}{N_b} \sum_{v=1}^{N_b} \sum_{i=1}^{k_1} p^{dd}(i |v) \log \frac{p^{dd}(i |v)}{1/R}
\end{equation}
\begin{equation}
\mathcal{L}_{k_1}^{ds} = - \frac{1}{N_b} \sum_{v=1}^{N_b} \sum_{j=1}^{k_1} p^{ds}(j |v) \log \frac{p^{ds}(j |v)}{1/R}.
\end{equation}
\end{subequations}

For each view, the loss is formulated as a weighted sum of the original neighborhood loss \(\mathcal{L}_{\Omega}\), the strict neighborhood loss \(\mathcal{L}_{k_1}\), and the extended neighborhood loss \(\mathcal{L}_{k_2}\). Specifically, the intra-view loss \(\mathcal{L}^{dd}\) and the cross-view loss \(\mathcal{L}^{ds}\) are, respectively, defined as
\begin{subequations}
\begin{equation}
\mathcal{L}_{c4}^{dd} = \mathcal{L}_{\Omega}^{dd} + \lambda_1 \mathcal{L}_{k_1}^{dd} + \lambda_2 \mathcal{L}_{k_2}^{dd}
\end{equation}
\begin{equation}
\mathcal{L}_{c4}^{ds} = \mathcal{L}_{\Omega}^{ds} + \lambda_1 \mathcal{L}_{k_1}^{ds} + \lambda_2 \mathcal{L}_{k_2}^{ds}
\end{equation}
\end{subequations}
where \(\lambda_1\) and \(\lambda_2\) are hyperparameters controlling the contribution of strict and extended neighborhood components. When using the satellite-view as the query,  \(\mathcal{L}_{c4}^{sd}\) and \(\mathcal{L}_{c4}^{ss}\) can be obtained in the same way. The total neighbor learning loss is computed as \(\mathcal{L}_{c4}=\mathcal{L}_{c4}^{dd}+\mathcal{L}_{c4}^{ds}+\mathcal{L}_{c4}^{sd}+\mathcal{L}_{c4}^{ss}\).

Finally, the overall loss of DMNIL is defined as
\begin{equation}\label{eq23}
\mathcal{L}_{total} = \mathcal{L}_{c1} + \mathcal{L}_{c3} + \mathcal{L}_{c4}.
\end{equation}

{\color{black} \remark{The ICEL module computes the association distribution between query samples and all candidate samples using a similarity measure $\mathcal{S}(\cdot, \cdot)$ and constructs multi-scale neighborhood structures through threshold-based and ranking-based selection. For a drone query, this includes a high-confidence distribution neighborhood $\Omega^{dd}$ and $\Omega^{ds}$, an extended neighborhood $\mathcal{N}_{k_2}^{dd}$ and $\mathcal{N}_{k_2}^{ds}$, and a strict neighborhood $\mathcal{N}_{k_1}^{dd}$ and $\mathcal{N}_{k_1}^{ds}$. Within the extended neighborhood, a normalized similarity distribution is computed and optimized by minimizing entropy. In the strict neighborhood, a conditional probability distribution between the query and its neighbors is established and the KL divergence to a uniform distribution is minimized. Satellite queries are processed in the same manner.}}

\subsection{Pseudo-Label Enhancement }
In self-supervised learning, the quality of pseudo-labels exerts a decisive influence on model performance. Inspired by \cite{yang2024shallow}, we propose a pseudo-label optimization strategy based on dynamic feature robustness. Specifically, we introduce small perturbations to the original feature representations and exploit the top-K neighbor consistency between the original and perturbed features to identify high-confidence samples while filtering out noisy instances. Then, a spatial smoothing operation based on the similarity matrix is applied to refine the pseudo-labels to enhance local label consistency.

Small Gaussian noise \(\boldsymbol{\epsilon}_d, \boldsymbol{\epsilon}_s \sim \mathcal{N}(\mathbf{0}, \sigma^2 \mathbf{I})\) is used in the initial satellite instances feature and the initial drone instances features to obtain perturbed features, where $\sigma$ denotes the standard deviation. It can be formulated as
\begin{equation} 
\mathbf{f}_i^{d{\boldsymbol{\epsilon}}} = \mathbf{f}_i^{d} + \boldsymbol{\epsilon}_d, \quad 
\mathbf{f}_j^{s{\boldsymbol{\epsilon}}} = \mathbf{f}_j^{s} + \boldsymbol{\epsilon}_s.
\end{equation}

Using the definition of the similarity score  shown in Eq.(\ref{eq16}),  two cross-view ranking lists can be obtained, which are denoted as $\mathcal{K}^{ds}=\{\mathcal{S}(\mathbf{f}_i^{d},\mathbf{f}_j^{s})\}$ and $\mathcal{K}^{ds}_{\boldsymbol{\epsilon}}=\{\mathcal{S}(\mathbf{f}_i^{d{\boldsymbol{\epsilon}}},\mathbf{f}_j^{s{\boldsymbol{\epsilon}}}))\}$, $i=1,\cdots,N$. The label of the $k$-th similar drone instance in two ranking lists can be denoted as \(y_{\mathbf{f}_i^d}[k]\) and \(y_{\mathbf{f}_i^{d{\boldsymbol{\epsilon}}}}[k]\). Next, the cross-view labels are associated with the intersection of two label sets to investigate collaborative ranking consistency. This process can be described as
\begin{equation}
    I_{\mathbf{f}_i^{d}} = \{y_{\mathbf{f}_i^d}[k] \cap y_{\mathbf{f}_i^{d{\boldsymbol{\epsilon}}}}[k]\}_{k=1}^N,
\end{equation}
where $I_{\mathbf{f}_i^{d}}$ records the samples with the same identity in instances. Consequently, we update the cross-view pseudo-label for $\mathbf{f}_j^s$ using the label with the highest frequency in $I_{\mathbf{f}_i^d}$, which reflects the most stable identity under both original and perturbed feature rankings. This final refined label is denoted as $\tilde{y}_{\mathbf{f}_j^{s}}$. Then, we convert the label list $\{\tilde{y}_{\mathbf{f}_j^{s}}\}$ into the form of one-hot code matrix $\tilde{\mathbf y} \in \mathbb{R}^{M \times N}$ by setting the column according to the refined labels to 1 and the rest to 0. 

In order to explore the intra-view initial and perturbed collaboration, the initial and perturbed similarity matrices \( \mathbf P^{ss} \) and \( \mathbf P^{ss}_{\boldsymbol{\epsilon}} \) are calculated, and their sum \(\mathbf  P \) is used to investigate intra-view ranking consistency. Then we keep the 5-max values of $\mathbf P$ in each row at 1 and set the rest to 0, acquiring the ranking relations. The process of intra-view ranking smoothing is formulated as $\mathbf y = \mathbf P\tilde{\mathbf y}$, where $\mathbf y\in \mathbb{R}^{M \times N}$ is the final refined cross-view label matrix of the satellite instance. In $\mathbf y$, the column number of the maximum value in each row is the refined label of the samples.
\vspace{-0.5cm}
\begin{table}[h]
\begin{center}
  \centering
  \caption{Comparisons between DMNIL and existing methods on University-1652}
  \label{tab:comparison}
\setlength{\tabcolsep}{3pt}
\resizebox{\columnwidth}{!}{ 
  \begin{tabular}{ccccccc}
  \hline \hline
   &&  \multicolumn{2}{c}{University-1652} \\ \cline{1-7}
   &  &  & \multicolumn{2}{c}{Drone$\rightarrow$Satellite} & \multicolumn{2}{c}{Satellite$\rightarrow$Drone} \\ \cline{4-7}
\multirow{-2}{*}{Model}  & \multirow{-2}{*}{Learning Type} & \multirow{-2}{*}{Parameters(M)} & {R@1} & {AP} & {R@1} & {AP} \\ \hline
  MuSe-Net\cite{wang2024multiple}            &Supervised& 50.47       & 74.48       & 77.83      & 88.02     & 75.10 \\
  MCCG\cite{shen2023mccg}                 &Supervised& 56.65       & 89.40       & 91.07      & 95.01     & 89.93 \\
  SDPL\cite{chen2024sdpl}                &Supervised& 42.56       & 90.16       & 91.64      & 93.58     & 89.45  \\
  Sample4Geo\cite{deuser2023sample4geo}   &Supervised& 87.57       & 92.65       & 93.81      & 95.14     & 91.39 \\
  SRLN\cite{lv2024direction}               &Supervised& 193.03      & 92.70       & 93.77      & 95.14     & 91.97 \\
  MEAN\cite{chen2024multi}         &Supervised &36.50 &93.55  &94.53 &{96.01}  & 92.08\\
  DAC\cite{xia2024enhancing}        &Supervised& 96.50      & 94.67       & 95.50      & 96.43     & 93.79 \\\hline
EM-CVGL\cite{li2024learning}   &Self-Supervised  & 2.25  & 70.29      & 74.93       & 79.03      & 61.03    \\
CDIKTNet\cite{chen2025limited} &Self-Supervised &91.52 & 83.30  &85.73 &{87.73} & 76.53\\
Wang\cite{wang2025coarse} &Self-Supervised &- &85.95 &90.33 &94.01 &82.66\\\hline
  DMNIL   &Self-Supervised& 28.59      & \textbf{90.17}       & \textbf{91.67}      & \textbf{95.01}     & \textbf{88.95}\\\hline\hline
\end{tabular}}
\end{center}
\end{table}
\vspace{-0.8cm}
\section{Experimental results}\label{results}
\subsection{Experimental Datasets and Evaluation Metrics}
To evaluate the proposed method, we conduct experiments on three benchmark datasets, including University-1652 \cite{zheng2020university}, SUES-200 \cite{zhu2023sues} and DenseUAV \cite{wang2024multiple}.

\textit{University-1652} contains multi-view images of 1,652 buildings across 72 universities worldwide, including drone, satellite, and ground-level views. The training set includes 701 buildings from 33 universities, and the test set covers 951 buildings from 39 geographically distinct universities. \textit{SUES-200} emphasizes altitude diversity, consisting of 200 locations (120 for training and 80 for testing). Each location provides one satellite image and four aerial images captured at different altitudes, covering urban, natural, and aquatic scenes. \textit{DenseUAV} focuses on low-altitude drone self-positioning, with over 27,000 drone–satellite images from 14 university campuses. It features dense sampling, multi-scale satellite imagery, and diverse temporal conditions, presenting challenges such as cross-view misalignment and spatial–temporal variations.

We evaluate the models using Recall@K (R@K) and Average Precision (AP), where R@K measures the proportion of correct matches among the Top-\(K\)results and AP evaluates the overall precision-recall performance. Model efficiency is assessed by the number of parameters to reflect portability under resource constraints. All comparisons are conducted using the best configurations for each method.

\begin{table*}[t]
  \centering
  \footnotesize
  \caption{Comparisons between DMNIL and existing methods on  SUES-200. ${\dagger}$ indicates results from self-supervised training on University-1652 and testing on SUES-200. ${\ddagger}$ denotes results from self-supervised training and evaluation on SUES-200. ${\ast}$~corresponds to results after one epoch self-supervised fine-tuning on SUES-200 based on the ${\dagger}$ pre-trained model
}
  \label{tab:comparison SUES-200-1}
  \setlength{\tabcolsep}{3pt}
   \resizebox{\textwidth}{!}{
  \begin{tabular}{ccccccccccc|cccccccc}
  \hline\hline
   &  \multicolumn{18}{c}{SUES-200} \\ \cline{1-19}
   &    & &\multicolumn{8}{c|}{Drone$\rightarrow$Satellite} &\multicolumn{8}{c}{Satellite$\rightarrow$Drone} \\ \cline{4-19} 
   &    & & \multicolumn{2}{c}{150m} & \multicolumn{2}{c}{200m} & \multicolumn{2}{c}{250m} & \multicolumn{2}{c|}{300m} & \multicolumn{2}{c}{150m} & \multicolumn{2}{c}{200m} & \multicolumn{2}{c}{250m} & \multicolumn{2}{c}{300m} \\ \cline{4-19} 
  \multirow{-3}{*}{Model} & \multirow{-3}{*}{Learning Type} & \multirow{-3}{*}{Parameters(M)}  & R@1 & AP & R@1 & AP & R@1 & AP & R@1 & AP  & R@1 & AP & R@1 & AP & R@1 & AP & R@1 & AP\\ \hline
  MCCG\cite{shen2023mccg}      &Supervised& 56.65  & 82.22 & 85.47 & 89.38 & 91.41 & 93.82 & 95.04 & 95.07 & 96.20 & 93.75 & 89.72 & 93.75 & 92.21 & 96.25 & 96.14 & 98.75 & 96.64 \\
  SDPL\cite{chen2024sdpl}     &Supervised& 42.56  & 82.95 & 85.82 & 92.73 & 94.07 & 96.05 & 96.69 & 97.83 & 98.05 & 93.75 & 83.75 & 96.25 & 92.42 & 97.50 & 95.65 & 96.25 & 96.17 \\

  SRLN\cite{lv2024direction}   &Supervised& 193.03 & 89.90 & 91.90 & 94.32 & 95.65 & 95.92 & 96.79 & 96.37 & 97.21 & 93.75 & 93.01 & 97.50 & 95.08 & 97.50 & 96.52 & 97.50 & 96.71 \\
  Sample4Geo\cite{deuser2023sample4geo}  &Supervised& 87.57  & 92.60 & 94.00 & 97.38 & 97.81 & 98.28 & 98.64 & 99.18 & 99.36 &{97.50} &93.63 &{98.75} &{96.70} &{98.75} &{98.28} &{98.75} &98.05 \\
  DAC\cite{xia2024enhancing}    &Supervised& 96.50 & 96.80  & 97.54 & 97.48 & 97.97 & 98.20 & 98.62 & 97.58 & 98.14 & 97.50 & 94.06  & 98.75 & 96.66 & 98.75 & 98.09 & 98.75 & 97.87 \\
  MEAN\cite{chen2024multi}       &Supervised& 36.50  & 95.50 & 96.46 & 98.38 & 98.72 & 98.95 & 99.17  & 99.52  & 99.63 & 97.50  & 94.75 & 100.00  & 97.09 & 100.00 & 98.28  & 100.00 & 99.21 \\\hline
    DMNIL  &Supervised& 28.59   & 94.00   & 95.21   & 97.83   & 98.88  &99.14  & 95.44  &99.25   &99.39    & 97.50   & 92.71   & 97.50   &  97.43  &100.00  & 98.98  &98.75   &98.25 \\\hline
  EM-CVGL\cite{li2024learning}   &Self-Supervised  & 2.25  & 55.23      & 60.80       & 60.95      & 61.03 & 68.10      & 72.62       & 74.42      & 78.20    & 73.75      &54.00       & 91.25      & 65.65  & 96.25      & 72.02       & 97.50      & 74.74  \\
  CDIKTNet\cite{chen2025limited}  &{Self-Supervised} &91.52 &{82.75} & {85.25} &{89.35} &{91.13} & {93.15} &{94.39} &{95.18} &{96.12} &{88.75}   &{80.33}   &{93.75} &{88.17}   &{95.00}   &{92.15}   &{98.75}    &{94.37} \\
  Wang\cite{wang2025coarse} &Self-Supervised &- &76.90 &84.95 &87.88 &92.60 &92.98 &95.66 &95.10 &96.92 &87.50 &74.81 &92.50 &87.15 &96.25 &91.20 &98.75 &94.52 \\\hline
  DMNIL$^{\dagger}$           &Self-Supervised& 28.59   & 82.60   & 85.60   & 90.48   & 92.28  &94.33  & 95.44  &96.78   &97.42  & 91.25   & 81.28   & 97.50   & 90.62  &98.75  & 94.74  &97.50   &96.17   \\
  DMNIL$^{\ddagger}$          &Self-Supervised& 28.59   & 87.00  & 89.20   & 93.98   & 95.11  &96.50  & 97.23  &97.23   &97.86  & 95.00   & 85.46   & 97.50   & 92.72  &\textbf{100.00}  & 95.43  &\textbf{100.00}   &96.39   \\
  DMNIL$^{\ast}$                    &Self-Supervised & 28.59  & \textbf{90.98} &\textbf{92.66}  & \textbf{95.03} & \textbf{96.14} &\textbf{97.45}  & \textbf{98.03}  & \textbf{98.23}  & \textbf{98.64} & \textbf{96.25} & \textbf{90.05}  & \textbf{97.50} & \textbf{94.80} & 98.75  &  \textbf{96.82}  & 98.75  & \textbf{97.33} \\ \hline\hline

\end{tabular}}
\end{table*}
\subsection{Implementation Details}
A \textit{ConvNeXt-Tiny} model, pre-trained on \textit{ImageNet}, serves as the backbone for feature extraction, with a newly added classifier module initialized via the Kaiming method. During both training and testing, all input images are resized to \(3\times 384\times 384\). We employ several data-augmentation techniques, including random cropping, random horizontal flipping, and random rotation. The batch size is set to 64. We use the DBSCAN \cite{ester1996density} to generate pseudo-labels, where the neighborhood radius is set to 0.40 for drone-view features and 0.30 for satellite-view features, and the minimum number of points required to form a cluster is 4. For optimization purposes, we use the SGD optimizer with an initial learning rate of 0.001 to train the model for 30 epochs in total. The weight \(w{''}\) and \(w{'}\) in Eq.(\ref{eq12}) are set to 0.7 and 0.3, respectively. The \(\gamma\) in Eq.(\ref{eq17}) is set to 0.9. The top-\( k_1 \) and top-\( k_2 \) are set to 10 and 20, respectively. The $\lambda_0$, $\lambda_1$, and $\lambda_2$ are set to 0.2, 0.01, and 0.1, respectively. All experiments are conducted under the PyTorch framework on an Ubuntu 22.04 system equipped with four NVIDIA RTX 4090 GPUs. 

\subsection{Comparison with State-of-the-art Methods}
To evaluate the effectiveness of DMNIL, we compare it with the state-of-the-art supervised methods MuSe-Net \cite{wang2024multiple}, MCCG\cite{shen2023mccg}, SDPL \cite{chen2024sdpl}, Sample4Geo \cite{deuser2023sample4geo}, SRLN \cite{lv2024direction}, MEAN \cite{chen2024multi} and DAC \cite{xia2024enhancing}, and the self-supervised methods EM-CVGL \cite{li2024learning}, CDIKTNet \cite{chen2025limited} and Wang \cite{wang2025coarse}. Notably, self-supervised solutions for DVGL remain limited, and the development of effective self-supervised methods in this domain is still in its early stage. This highlights our contributions to the advancement of self-supervised learning for DVGL.


\textit{Results on University-1652}. As shown in Table~\ref{tab:comparison}, DMNIL demonstrates advantages over existing self-supervised methods. It achieves R@1/AP scores of 90.17\%/91.67\% and 95.01\%/88.95\% on the Drone$\rightarrow$Satellite and Satellite$\rightarrow$Drone settings, respectively. Compared with EM-CVGL, DMNIL improves R@1 and AP by about 20\%/17\% and 16\%/28\% in the two settings. Even against CDIKTNet, which benefits from paired data for self-supervised fine-tuning, DMNIL still achieves superior results. Unlike clustering-based methods that rely on explicit appearance transformations, DMNIL implicitly captures cross-view variations through neighborhood-guided consistency and mutual information constraints. As a result, DMNIL improvements of 4.22\% / 1.00\% in R@1 and 1.34\% / 6.29\% in the two settings. Moreover, DMNIL surpasses several supervised methods such as MuSe-Net, MCCG, and SDPL.

\textit{Results on SUES-200}.
As shown in Table \ref{tab:comparison SUES-200-1}, when directly transferring the pre-trained model from the University-1652 dataset to the SUES-200 dataset without any re-training, DMNIL consistently outperforms all existing self-supervised methods and achieves competitive performance compared with supervised methods such as SDPL. DMNIL outperforms some supervised methods such as MCCG and SDPL. Notably, when further fine-tuning the transferred model for only one epoch, we observe additional improvements that surpass the supervised SRLN. These results clearly demonstrate the effectiveness of DMNIL in self-supervised learning, transfer learning, and efficient fine-tuning.

\textit{Results on Cross-Domain Generalization Performance}.
As shown in Table \ref{tab:comparison University-SUES}, DMNIL consistently outperforms existing self-supervised methods in cross-domain setting. Specifically, in the Drone $\rightarrow$ Satellite setting, our model achieves better performance than EM-CVGL at all altitudes. At 150 meters, R@1 and AP increase by approximately 22\% and 20\%. At 200 meters, the lifts from DMNIL are around 19\% and 16\%. At 250 meters, they are nearly 14\% and 12\%. Even at 300 meters, DMNIL still outperforms about 11\% and 9\%. In the Satellite $\rightarrow$ Drone scenario, similar improvements are observed. At 150 meters, R@1 and AP increase by approximately 18\% and 25\%, respectively. At 200 meters, the improvements are around 10\% and 20\%. Moving further to 250 meters, they are lifted by 6\% and 13\%. Finally, at 300 meters, the lifts remain considerable with R@1 increasing by 3\% and AP increasing by 10\%. Remarkably, DMNIL even surpasses all existing state-of-the-art supervised methods in terms of cross-domain generalization. These results confirm the robustness and strong cross-domain generalization capability of DMNIL at varying altitudes. 

\begin{table*}[t]
  \centering
  \footnotesize
\caption{Comparisons between DMNIL and existing methods in cross-domain evaluation}
\label{tab:comparison University-SUES}
  \setlength{\tabcolsep}{3pt}
   \resizebox{\textwidth}{!}{
  \begin{tabular}{ccccccccccc|cccccccc}
  \hline\hline
   &  \multicolumn{18}{c}{University-1652$\rightarrow$SUES-200} \\ \cline{1-19}
   &    & &\multicolumn{8}{c|}{Drone$\rightarrow$Satellite} &\multicolumn{8}{c}{Satellite$\rightarrow$Drone} \\ \cline{4-19} 
   &    & & \multicolumn{2}{c}{150m} & \multicolumn{2}{c}{200m} & \multicolumn{2}{c}{250m} & \multicolumn{2}{c|}{300m} & \multicolumn{2}{c}{150m} & \multicolumn{2}{c}{200m} & \multicolumn{2}{c}{250m} & \multicolumn{2}{c}{300m} \\ \cline{4-19} 
  \multirow{-3}{*}{Model} & \multirow{-3}{*}{Learning Type} & \multirow{-3}{*}{Parameters(M)}  & R@1 & AP & R@1 & AP & R@1 & AP & R@1 & AP & R@1 & AP & R@1 & AP & R@1 & AP & R@1 & AP \\ \hline
  MCCG\cite{shen2023mccg}      &Supervised& 56.65  & 57.62 & 62.80 & 66.83 & 71.60 & 74.25 & 78.35 & 82.55 & 85.27  & 61.25 & 53.51 & 82.50 & 67.06 & 81.25 & 74.99 & 87.50 & 80.20 \\
 
  Sample4Geo\cite{deuser2023sample4geo}  &Supervised& 87.57  & 70.05 & 74.93 & 80.68 & 83.90 & 87.35 & 89.72 & 90.03 & 91.91 & 83.75 & 73.83 & 91.25 & 83.42 & 93.75 & 89.07 & 93.75 & 90.66  \\
  CAMP\cite{wu2024camp}  &Supervised& 91.40  & 78.90 & 82.38 & 86.83  & 89.28 & 91.95 & 93.63 & 95.68  & 96.65 & 87.50 & 78.98 & 95.00 & 87.05  & 95.00 & 91.05 & 96.25 & 93.44   \\
  DAC\cite{xia2024enhancing}    &Supervised& 96.50 & 76.65 & 80.56 & 86.45 & 89.00 & 92.95  & 94.18 & 94.53 & 95.45 & 87.50 & 79.87 & 96.25 & 88.98  & 95.00 & 92.81 & 96.25 & 94.00  \\
  MEAN\cite{chen2024multi}       &Supervised& 36.50  & 81.73 & 87.72  & 89.05  & 91.00  & 92.13  & 93.60  & 94.63  & 95.76 & 91.25   & 81.50   & 96.25   &  89.55  & 95.00 & 92.36 & 96.25  & 94.32 \\\hline
  
  EM-CVGL\cite{li2024learning} &Self-Supervised  & 2.25  & 60.03  &65.69       & 71.50      & 76.17  & 80.35  & 83.85       & 85.93      & 88.34  & 73.75  &56.99       & 87.50      & 70.62  & 92.50  & 81.18       & 95.00      & 86.04  \\
DMNIL    &Self-Supervised& 28.59   & \textbf{82.60}   & \textbf{85.60}   & \textbf{90.48}   & \textbf{92.28}  &\textbf{94.33}  & \textbf{95.44}  &\textbf{96.78}   &\textbf{97.42} &\textbf{91.25}  & \textbf{81.28}   & \textbf{97.50}   & \textbf{90.62}  &\textbf{98.75}  & \textbf{94.74}  &\textbf{97.50}   &\textbf{96.17}
  
   \\ \hline\hline

\end{tabular}}
\end{table*}

\textit{Results on DenseUAV}.
The DenseUAV dataset presents a highly challenging scenario due to its densely sampled nature, where neighboring images exhibit high visual similarity. It often leads to severe confusion in DVGL. In addition, the dataset includes satellite images captured at different times and drone images from multiple altitudes, which further increases intra-domain variability and task complexity. As shown in Table \ref{tab:DenseUAV}, in the fully self-supervised setting, DMNIL achieves 73.79\% R@1 and 90.69\% R@5 across all altitudes, outperforming existing self-supervised methods. It is noteworthy that the two representative self-supervised methods consistently yield poor performance on DenseUAV, due to their limitations in handling densely sampled and spatio-temporal diverse scenarios. These results demonstrate the effectiveness of DMNIL in addressing feature confusion caused by spatial continuity and appearance shifts induced by temporal variation.

\subsection{Ablation Studies}
To thoroughly evaluate the effectiveness of our proposed method, we perform an ablation study on the University-1652 dataset. We systematically assess the impact of each individual module and investigate their contributions under different configurations. The results are presented in Table \ref{Ablation Studies}.

\textit{ConvNeXt-Tiny Backbone}.
The backbone refers to the untrained ConvNeXt-Tiny network used as a direct feature extractor. As shown in Table \ref{Ablation Studies}, its performance is significantly limited, and obtains an R@1 of only {9.55\%} and {35.24\%}, and an AP of {11.88\%} and {13.20\%} for the {Drone} $\rightarrow$ {Satellite} and {Satellite} $\rightarrow$ {Drone}, respectively. These results indicate that the backbone alone lacks the ability to extract discriminative cross-view features.

\begin{table}[h]
  \caption{Comparisons between DMNIL and existing methods on DenseUAV}
  \label{tab:DenseUAV}
  \footnotesize
\setlength{\tabcolsep}{3pt}
\resizebox{\columnwidth}{!}{ 
  \begin{tabular}{ccccccccccc}
  \hline \hline
   &  \multicolumn{7}{c}{DenseUAV} \\ \cline{1-11}
   &    && \multicolumn{2}{c}{All height}& \multicolumn{2}{c}{80m} &\multicolumn{2}{c}{90m}& \multicolumn{2}{c}{100m}  \\ \cline{4-11}
\multirow{-2}{*}{Model} & \multirow{-2}{*}{Learning Type} & \multirow{-2}{*}{Parameters(M)} & {R@1} & {R@5} & {R@1} & {R@5} & {R@1} & {R@5} & {R@1} & {R@5}\\ \hline
  ConvNext-Tiny\cite{DenseUAV}       &Supervised& 30.10  & 60.23 & 81.94 & -  & - & -  & -& -  & -\\
  Sample4Geo\cite{deuser2023sample4geo}    &Supervised& 87.57& 80.57 &96.53  &80.57&96.53 &86.87&98.71 &91.38& 99.74 \\
  CAMP\cite{wu2024camp}                 &Supervised& 91.40  & 88.72 & 97.76 &76.45  &96.01 &84.81  &99.23 & 89.19  &99.23 \\
  DAC\cite{xia2024enhancing}            &Supervised&96.50 & 84.47 & 96.53 &80.92 &97.04  &85.32 &98.71 &85.46 &99.10\\
  MEAN\cite{chen2024multi}             &Supervised& 36.50   &90.18 &97.86        &78.76 &97.04 &85.33 &99.10 &89.32 &98.97 \\\hline
  EM-CVGL\cite{li2024learning}  &Self-Supervised  & 2.25  & 18.15  & 50.97       & 18.02      &50.36   & 17.76  & 50.19       & 17.50      & 49.98    \\
  CDIKTNet\cite{chen2025limited} &{Self-Supervised} &91.52 &{22.57}   &{40.24}   &{17.25} &{40.15}   &{23.04}   &{48.39}   &{25.48}    &{52.38}\\
   DMNIL    &Self-Supervised& 28.59        & \textbf{73.79} & \textbf{90.69}      &\textbf{73.10} &\textbf{92.92}    &\textbf{74.65} &\textbf{95.37}   &\textbf{75.03} &\textbf{95.50}\\ 
  \hline\hline
\end{tabular}}
\end{table}
\textit{Effectiveness of Dual-Path Contrastive Learning Baseline}.
After incorporating the baseline, R@1 improves to 71.06\% and 81.31\%, while AP reaches 74.91\% and 65.26\%, respectively. Although the baseline primarily learns intra-view feature discrimination, the training process also facilitates a certain degree of alignment between drone and satellite representations. This leads to the formation of a more consistent and unified feature representation space, which serves as a solid foundation for subsequent cross-view learning.

\textit{Effectiveness of Dynamic Hierarchical Memory Learning}.
Compared with baseline, incorporation of DHML leads to performance improvements in both settings. Specifically, R@1 increases by 12.97\% and 9.13\%, and AP improves by 11.5\% and 17.48\%, respectively. These gains can be attributed to the joint modeling of long-term and short-term feature dependencies introduced by DHML, which enhances intra-view feature representation and discrimination.
\begin{table}[t]
  \caption{Influence of each component on performance of DMNIL}
  \setlength{\tabcolsep}{3pt}
\resizebox{\columnwidth}{!}{ 
  \begin{tabular}{ccccccccc}
  \hline\hline
  \multicolumn{5}{c}{\multirow{2}{*}{Setting}} & \multicolumn{4}{c}{University-1652}                                       \\ \cline{6-9} 
  \multicolumn{5}{c}{}                         & \multicolumn{2}{c}{Drone$\rightarrow$Satellite} & \multicolumn{2}{c}{Satellite$\rightarrow$Drone} \\ \hline
   Backbone          &Baseline         & DHML      &ICEL       & PLE  & R@1    & AP       & R@1    & AP               \\ \hline
\checkmark &         &                 &       &             & 9.55  & 11.88    & 35.24  & 13.20 \\
 \checkmark& \checkmark        &      &       &             & 71.06  & 74.91    & 81.31  & 65.26 \\
 \checkmark& \checkmark  &\checkmark    &      &            & 84.03  & 86.41    & 90.44  & 82.74  \\
  \checkmark&\checkmark  &    &\checkmark      &         & 74.48  & 77.96    & 89.87  & 74.37     \\
\checkmark &  \checkmark & \checkmark      & \checkmark     & & 86.13    & 88.31  & 94.01 & 82.97 \\
\checkmark & \checkmark  & \checkmark      & \checkmark     & \checkmark &\textbf{90.17}    & \textbf{91.67} & \textbf{95.01}  & \textbf{88.95}            \\ \hline\hline
\label{Ablation Studies}
\end{tabular}}
\vspace{-1.8em}
\end{table}

\textit{Effectiveness of Information Consistency Evolution Learning}.
{\color{black}Although DHML enhances intra-view feature representation and discrimination, it still exhibits notable limitations in learning cross-view information. To address this, ICEL is proposed to learn cross-view dependencies. By enabling bidirectional information interaction between different views, ICEL facilitates cross-view feature refinement. As a result, R@1 improves to 86.13\% and 94.01\%, while AP reaches 88.31\% and 82.97\%. These results validate the effectiveness of ICEL in enhancing cross-view consistency for DVGL.}

\textit{Pseudo-Label Enhancement Refining Feature Learning with Robust Labels}.
To mitigate pseudo-label noise caused by clustering, we apply PLE to improve label reliability. After incorporating PLE, our final model achieves state-of-the-art performance, with R@1 increasing to 90.17\% and 95.01\%, and AP reaching 91.67\% and 88.95\%. This confirms that refining pseudo-label quality enhances the final model's effectiveness in DVGL.


\begin{figure*}[ht]
  \includegraphics[width=7.0in]{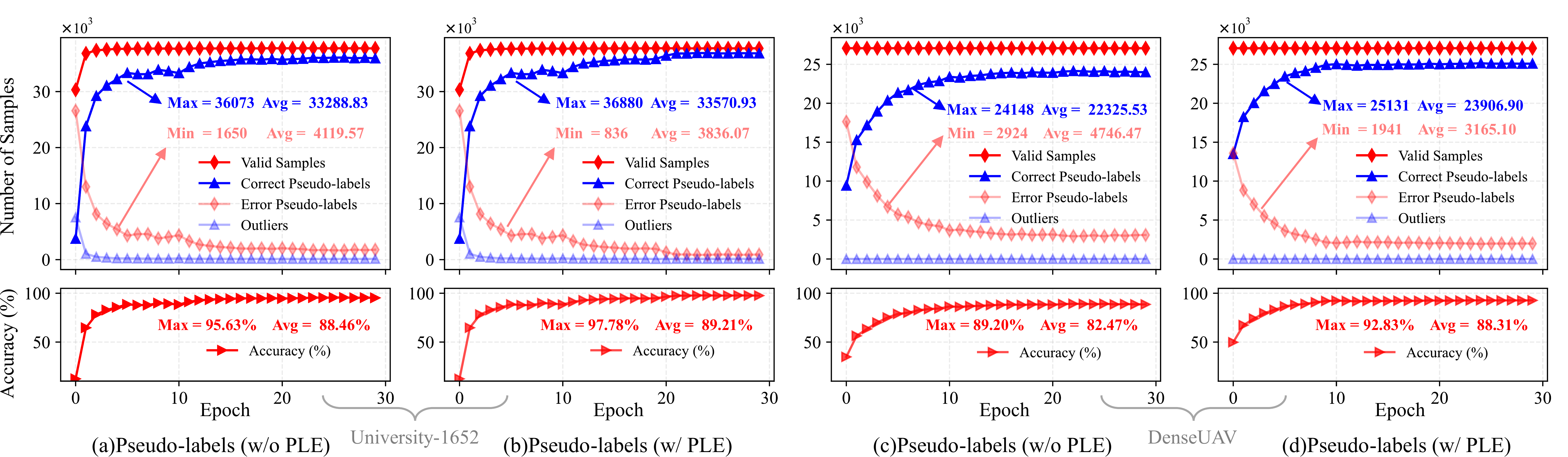}
  \caption{\textbf{Accuracy of Pseudo-Labels}. (a-b): Evolution of pseudo-label quality during training on the University-1652. (c-d): Evolution of pseudo-label quality during training on the DenseUAV datasets. ``Pseudo-labels (w/o PLE)" and ``Pseudo-labels (w/ PLE)" denote models trained without and with the proposed pseudo-label enhancement (PLE) strategy, respectively. The upper plots show the number of valid samples, correct and error pseudo-labels, and outliers obtained at each epoch, while the lower plots present the corresponding pseudo-label accuracy. The abbreviations ``Max," ``Min," and ``Avg" in the table denote the maximum, minimum, and average values, respectively.}
  \label{label}
  \end{figure*}
  
\begin{table}[h]
\caption{Influence of DBSCAN EPS parameter applied to drone-view on University-1652.}
\label{tab:eps_drone}
\resizebox{\columnwidth}{!}{ 
\begin{tabular}{cccccccc}
\hline\hline
 &\multirow{2}{*}{Eps} & \multirow{2}{*}{Number of Clusters} & \multirow{2}{*}{Number of Outliers} & \multicolumn{2}{c}{Drone$\to$Satellite} & \multicolumn{2}{c}{Satellite$\to$Drone} \\
\cline{5-8}
  &&  &  & R@1 & AP & R@1 & AP \\
\hline
 \multirow{5}{*}{\rotatebox{90}{Drone}}&0.1 & 384 & 35390 & 22.23 & 26.63 & 60.91 & 14.92 \\
 &0.2 & 1249 & 26498 & 54.60 & 59.40 & 90.87 & 52.64 \\
 &0.3 & 1785 & 15777 & 77.66 & 80.95 & 91.44 & 72.29 \\
 &0.4 & 1656 & 7535 & \textbf{90.17} & \textbf{91.67} & \textbf{95.01} & \textbf{88.95} \\
 &0.5 & 1100 & 2319 & 89.86 & 91.44 & 94.43 & 88.88 \\
\hline\hline
\end{tabular}}
\end{table}

\begin{table}[ht]
\caption{Influence of DBSCAN EPS parameter applied to satellite-view on University-1652}
\resizebox{\columnwidth}{!}{ 
\label{tab:eps_satellite}
\begin{tabular}{cccccccc}
\hline\hline
&\multirow{2}{*}{EPS} & \multirow{2}{*}{Number of Clusters} & \multirow{2}{*}{Number of Outliers} & \multicolumn{2}{c}{Drone$\to$Satellite} & \multicolumn{2}{c}{Satellite$\to$Drone} \\
\cline{5-8}
 &&  &  & R@1 & AP & R@1 & AP \\
\hline
\multirow{5}{*}{\rotatebox{90}{Satellite}}  &0.1 & 831 & 8 & 89.29 & 90.89 & 94.57 & 87.74 \\
&0.2 & 827 & 8 & 89.77 & 91.32 & 94.43 & 88.78 \\
&0.3 & 827 & 8 & \textbf{90.17} & \textbf{91.67} & \textbf{95.01} & \textbf{88.95} \\
&0.4 & 821 & 8 & 88.72 & 90.41 & 94.43 & 87.28 \\
&0.5 & 811 & 0 & 89.64 & 91.18 & 94.72 & 88.10 \\
\hline\hline
\end{tabular}}
\end{table}

\begin{table}[ht]
\caption{
Performance evaluation of different clustering algorithms on University-1652}
\label{tab:clustering_results}
\resizebox{\columnwidth}{!}{ 
\begin{tabular}{lcccccccc}
\hline\hline
\multirow{2}{*}{Algorithm} & 
\multicolumn{2}{c}{Number of Clusters} & 
\multicolumn{2}{c}{Number of Outliers} & 
\multicolumn{2}{c}{Drone$\to$Satellite} & 
\multicolumn{2}{c}{Satellite$\to$Drone} \\
\cmidrule(lr){2-3} \cmidrule(lr){4-5} \cmidrule(lr){6-7} \cmidrule(lr){8-9}
& Drone & Satellite & Drone & Satellite & R@1 & AP & R@1 & AP \\
\midrule
HDBSCAN\cite{mcinnes2017hdbscan} & 1260 & 818 & 5563 & 0 & 1.22 & 2.28 & 0.85 & 0.99 \\
Agglomerative Clustering\cite{ward1963hierarchical} & 28697 & 840 & 0 & 0 & 29.54 & 29.54 & 67.18 & 29.59 \\
DBSCAN\cite{ester1996density} & 1656 & 827 & 7535 & 8 & \textbf{90.17} & \textbf{91.67} & \textbf{95.01} & \textbf{88.95} \\
\hline\hline
\end{tabular}}
\end{table}

\subsection{Further Analysis}

{\color{black} \textit{Clustering Analysis.} We further conducted a sensitivity analysis of the DBSCAN parameters and performance comparisons of different clustering algorithms. As shown in Tables \ref{tab:eps_drone} and \ref{tab:eps_satellite}, the parameter EPS influences both the number of clusters and outliers that directly impact  model performance. For the drone-view, as EPS increases from 0.1 to 0.5, the number of clusters initially rises and then falls, while outliers decrease sharply. The best performance is achieved at EPS = 0.4. When EPS becomes too large, dissimilar samples are incorrectly merged, which leads to a slight performance drop. For the satellite-view, results remain relatively stable, with the best performance at EPS = 0.3. As shown in Table \ref{tab:clustering_results}, among three different clustering algorithms, HDBSCAN requires no manual tuning but yields many small clusters and lower accuracy. Agglomerative Clustering groups nearly all samples but performs suboptimally. In contrast, DBSCAN filters noisy samples and form compact, discriminative clusters, which achieves the best performance.}

{\color{black} \textit{Analysis of the Accuracy of
Pseudo-Labels During the Training Stage.} We further investigate the variations of pseudo-labels during training on the University-1652 and DenseUAV datasets, as shown in Fig. \ref{label}. On University-1652, the number of valid samples rapidly increases and stabilizes at about $33\times10^3$. The pseudo-label accuracy reaches 95.63\% at its peak and maintains an average of 88.46\%. With the proposed PLE module, both the valid samples and accuracy are further improved, while the number of erroneous pseudo-labels gradually decreases during training. On DenseUAV, a similar trend can be observed. Without PLE, valid pseudo-labels stabilize around $22\times10^3$ and the accuracy reaches 89.20\% with an average of 82.47\%. After applying PLE, valid samples increase to $23\times10^3$ and the accuracy improves to 92.83\% with an average of 88.31\%. These results confirm that PLE effectively improves pseudo-label quality.}

{\color{black} \textit{Analysis of Perturbation Effect}. To further evaluate the influence of perturbation in the PLE, we compare the performance under settings with and without perturbation. As shown in Table~\ref{tab:perturbation_results}, the model trained without perturbation achieves stable but slightly lower retrieval accuracy. Introducing moderate Gaussian perturbation (\(\sigma = 0.05 \sim 0.15\)) consistently improves the results in both retrieval directions, indicating that the perturbation facilitates smoother feature alignment and enhances local consistency during training. The best performance is achieved at \(\sigma = 0.10\), which provides an optimal trade-off between feature diversity and stability. In contrast, an excessive perturbation level (\(\sigma = 0.20\)) introduces noise that disturbs the learned representations, leading to performance degradation.}

\begin{table}[h]
  \caption{Performance comparison with and without perturbation.}
    \tiny
  \centering
  \resizebox{\columnwidth}{!}{ 
  \begin{tabular}{cccccc}
  \hline\hline
    \multirow{2}{*}{Setting} & \multirow{2}{*}{$\sigma$} & \multicolumn{2}{c}{Drone$\to$Satellite} & \multicolumn{2}{c}{Satellite$\to$Drone} \\
    \cline{3-6}
     & & R@1 & AP & R@1 & AP \\\hline

    w/o Perturbation & - & 89.49 & 91.10 & 94.15 & 87.41 \\\hline
    \multirow{4}{*}{w/ Perturbation} & 0.05 & 90.17 & 91.67 & 95.01 & \textbf{88.95} \\
    & 0.10 & \textbf{90.45} & \textbf{91.94} & \textbf{95.15} & 88.85 \\
    & 0.15 & 90.37 & 91.92 & 95.15 & 88.74 \\
    & 0.20 & 89.20 & 90.89 & 94.02 & 86.00 \\\hline\hline
  \end{tabular}}
  \label{tab:perturbation_results}
\end{table}

\begin{figure}[h]
\centering
  \includegraphics[width=3.3in]{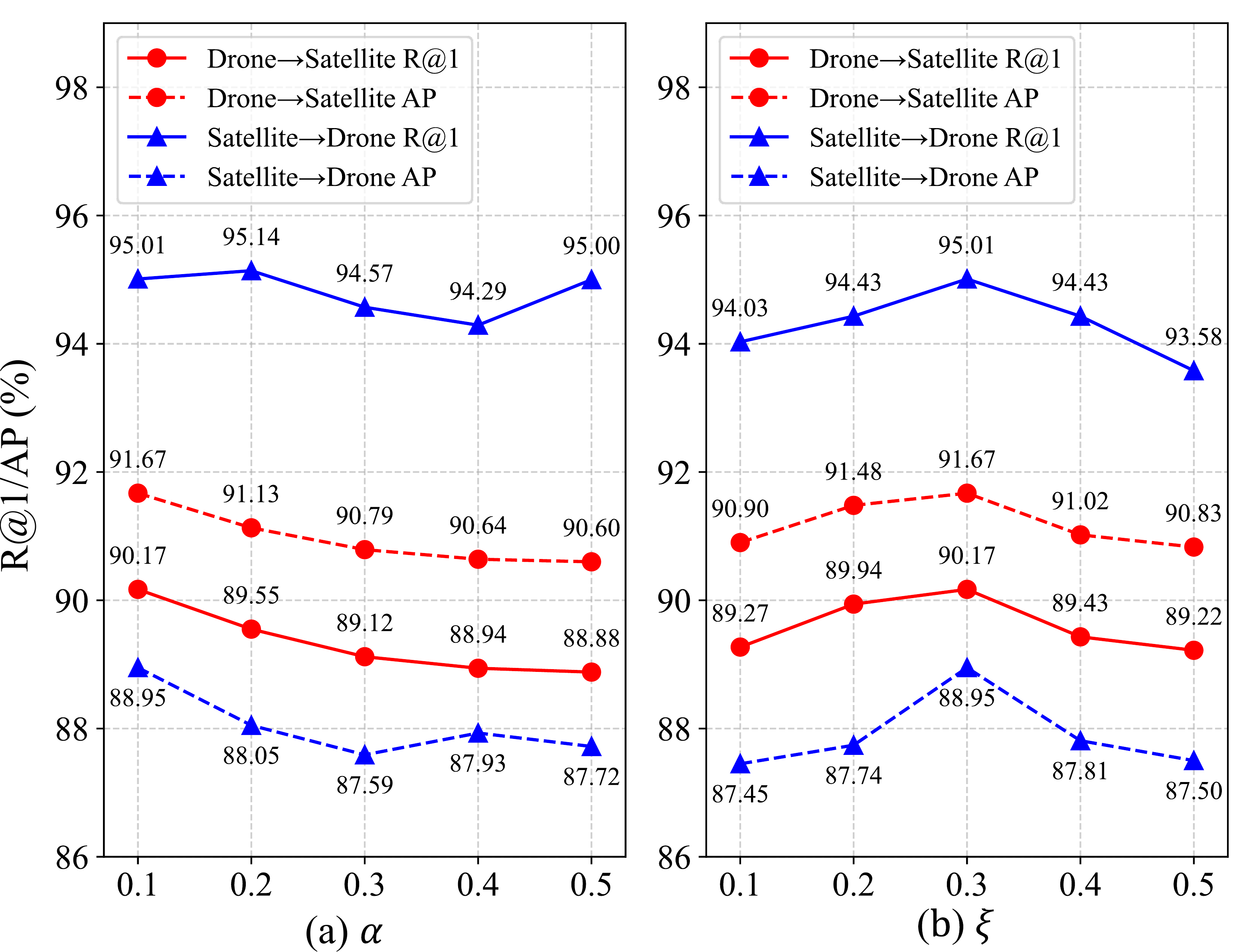}
  \caption{\textbf{Performance influence of momentum factors}. (a) Performance influence of the momentum factor $\alpha$ in the baseline  on the University-1652. (b) Performance influence of the momentum factor $\xi$ in the DHML module on the University-1652.}
  \label{momentum}
  \end{figure}

{\color{black} \textit{Hyper-parameter Analysis for $\alpha$ and $\xi$.} The influence of hyper-parameters $\alpha$ in Eq.(\ref{eq4}) and $\xi$ in Eq.(\ref{eq8}) are shown in Fig.~\ref{momentum}. In the baseline, which learns from cluster-level representations, $\alpha$ is set to 0.1, as adopted in previous works \cite{dai2022cluster,adca}. When the momentum factor increases in Eq.~(4), the update of the current query instance feature becomes slower, which weakens its adaptability to new samples and feature variations. In the DHML module, historical consistency is emphasized by maintaining a 7:3 ratio between historical and current features. This design balances stability and adaptability, which leads to the best performance when $\xi = 0.3$. This discrepancy arises from the fundamentally different motivations of momentum factor in the two designs. The baseline requires fast and responsive feature updates to track the evolving cluster assignments. In contrast, DHML aims to accumulate stable historical information and suppress noisy in early stage.}

\begin{figure}[h]
\centering
  \includegraphics[width=3.3in]{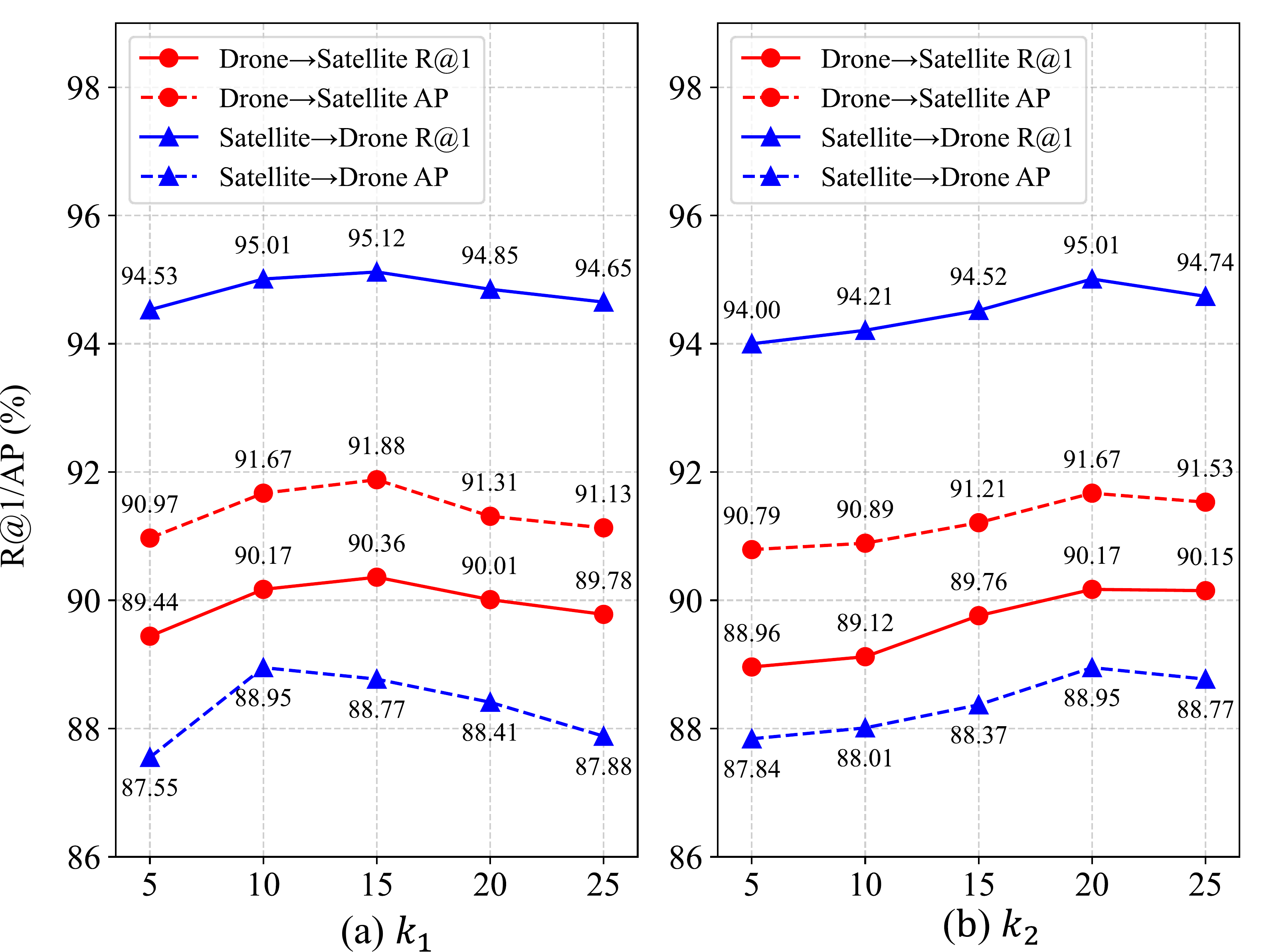}
  \caption{\textbf{Performance influence of the parameters $k_1$ and $k_2$ }.(a) Performance influence of the $k_1$ in the ICEL module on the University-1652 dataset. (b) Performance influence of the parameter $k_2$ in the ICEL module on the University-1652 dataset.}
  \label{topk}
  \end{figure}

{\color{black} \textit{Hyper-parameter Analysis for \(k_1\) and \(k_2\).} As shown in Fig.~\ref{topk}, the parameter \(k_1\) controls the size of a strictly filtered, high-confidence neighborhood, while \(k_2\) defines an expanded neighborhood that captures additional informative samples (\(k_2 > k_1\)). When \(k_1\) is too small, the neighborhood lacks sufficient diversity and informative positives, whereas a large \(k_1\) introduces low-confidence samples and noise. Similarly, increasing \(k_2\) enhances cross-view contextual relations, but excessive expansion may include irrelevant neighbors and lead to slight degradation. In this work, \(k_1\) and \(k_2\) are set to 10 and 20, respectively. It is noted that slightly adjusting these parameters may lead to further performance improvements.} 

{\color{black} \textit{Analysis of Robustness}. To further evaluate the robustness of DMNIL, the analysis considered sparsity with respect to both features and samples. For feature sparsity, we randomly masked three black patches on each test image to simulate missing visual information and reduce cross-view overlap. The patches were allowed to overlap. The ratio of the masked area to the total area was controlled by a parameter $A_s$. For sample sparsity, since each location in the original test set contains only one satellite image but multiple drone images, we controlled the number of available drone images per location to simulate varying degrees of sample sparsity. The corresponding results are reported in Table ~\ref{feature sparsity} and Table ~\ref{sample sparsity}.}


{\color{black} For {feature sparsity}, as shown in Table~\ref{feature sparsity}, when increasing the occluded area ratio $A_s$ from 0.1 to 0.5, both R@1 and AP of DMNIL decrease gradually but remain consistently higher than those of the baseline. Even at $A_s = 0.5$, where half of the visual content is randomly masked, DMNIL still achieves {36.88\% R@1} and {41.63\% AP}, outperforming the baseline by a considerable margin. For {sample sparsity}, as shown in Table~\ref{sample sparsity}, the performance of both methods declines as the number of available drone images per location decreases from 20 to 1. Nevertheless, DMNIL consistently achieves higher retrieval accuracy than the baseline, reaching {91.16\% R@1} with only a single drone image. The relatively small performance gap under reduced samples verifies that DMNIL generalizes well even when the cross-view matching pairs are extremely limited.}

\begin{table}[ht]
\caption{Comparison of DMNIL with baseline under feature sparsity process on drone$\rightarrow$satellite of University-1652}
\label{feature sparsity}
\footnotesize
\resizebox{\columnwidth}{!}{ 
\begin{tabular}{ccccc}
\hline\hline
\multirow{2}{*}{As (\%)}     & \multicolumn{2}{c}{DMNIL} & \multicolumn{2}{c}{Baseline} \\
\cline{2-5} 
  & R@1      & AP       & R@1     & AP      \\
\hline
0.1     &\textbf{{84.69$_{-5.480}$}} & \textbf{{86.99$_{-4.680}$}} & 64.67$_{-6.090}$ & 69.13$_{-5.780}$ \\
0.2      &\textbf{ {72.89$_{-17.28}$}} & \textbf{{76.34$_{-15.33}$}} & 52.54$_{-18.52}$ & 57.70$_{-17.21}$ \\
0.3       &\textbf{{58.89$_{-31.28}$}} &\textbf{{63.11$_{-28.56}$}} & 40.12$_{-30.94}$ & 45.22$_{-29.69}$ \\
0.4      &\textbf{{45.27$_{-44.90}$}} &\textbf{{49.98$_{-41.69}$}} & 30.22$_{-40.84}$ & 35.07$_{-39.84}$ \\
0.5      &\textbf{{36.88$_{-54.29}$}} &\textbf{{41.63$_{-50.04}$}} & 23.76$_{-47.30}$ & 28.31$_{-46.60}$ \\
\hline\hline
\end{tabular}}
\end{table}

\begin{table}[ht]
\caption{Comparison of DMNIL with baseline under sample sparsity process on satellite$\rightarrow$drone of University-1652}
\label{sample sparsity}
\footnotesize
\resizebox{\columnwidth}{!}{ 
\begin{tabular}{ccccc}
\hline\hline
\multirow{2}{*}{Number of Drone}     & \multicolumn{2}{c}{DMNIL} & \multicolumn{2}{c}{Baseline} \\
\cline{2-5} 
  & R@1      & AP       & R@1     & AP      \\
\hline
1     & \textbf{{91.16$_{-3.990}$}} & \textbf{{92.44$_{+3.590}$}} & {66.61$_{-14.70}$} & {70.52$_{+5.260}$} \\
5     & \textbf{{93.56$_{-1.590}$}} & \textbf{{89.07$_{+0.220}$}} & {76.03$_{-5.280}$} & {66.68$_{+1.420}$} \\
10    & \textbf{{93.72$_{-1.430}$}} & \textbf{{89.11$_{+0.260}$}} & {77.74$_{-3.570}$} & {65.60$_{+0.340}$} \\
15    & \textbf{{94.29$_{-0.860}$}} & \textbf{{89.15$_{+0.300}$}} & {78.60$_{-2.710}$} & {65.58$_{+0.320}$} \\
20    & \textbf{{95.10$_{-0.050}$}} & \textbf{{89.20$_{+0.350}$}} & {79.60$_{-1.710}$} & {65.43$_{+0.170}$} \\
\hline\hline
\end{tabular}}
\end{table}

\begin{table}[ht]
  \caption{Different configurations of learning type on University-1652}
  \label{tab:learning configurations1}
  \footnotesize
\resizebox{\columnwidth}{!}{ 
\begin{tabular}{ccccccc}
  \hline \hline
   &&  \multicolumn{2}{c}{University-1652} \\ \cline{1-7}
   &  &  & \multicolumn{2}{c}{Drone$\rightarrow$Satellite} & \multicolumn{2}{c}{Satellite$\rightarrow$Drone} \\ \cline{4-7}
\multirow{-2}{*}{Model}  & \multirow{-2}{*}{Learning Type} & \multirow{-2}{*}{Parameters(M)} & {R@1} & {AP} & {R@1} & {AP} \\ \hline

 Backbone&Supervised& 28.59      &{90.12}       &{91.96}      &{94.92}     &{89.72}\\\hline
   \multirow{2}{*}{DMNIL} &Self-Supervised& 28.59      & {90.17}       & {91.67}      & {95.01}     & {88.95}\\
  &Supervised& 28.59      & \textbf{92.02}       & \textbf{93.36}      & \textbf{95.57}     & \textbf{91.27} \\ \hline
  \hline
\end{tabular}}
\end{table}

\begin{table*}[ht]
  \footnotesize
  \caption{Different configurations of learning type on SUES-200}
  \label{tab:learning configurations2}
  \setlength{\tabcolsep}{3pt}
   \resizebox{\textwidth}{!}{
  \begin{tabular}{ccccccccccc|cccccccc}
  \hline\hline
   &  \multicolumn{18}{c}{SUES-200} \\ \cline{1-19}
   &    & &\multicolumn{8}{c|}{Drone$\rightarrow$Satellite} &\multicolumn{8}{c}{Drone$\rightarrow$Satellite} \\ \cline{4-19} 
   &    & & \multicolumn{2}{c}{150m} & \multicolumn{2}{c}{200m} & \multicolumn{2}{c}{250m} & \multicolumn{2}{c|}{300m} & \multicolumn{2}{c}{150m} & \multicolumn{2}{c}{200m} & \multicolumn{2}{c}{250m} & \multicolumn{2}{c}{300m} \\ \cline{4-19} 
  \multirow{-3}{*}{Model} & \multirow{-3}{*}{Learning Type} & \multirow{-3}{*}{Parameters(M)}  & R@1 & AP & R@1 & AP & R@1 & AP & R@1 & AP  & R@1 & AP & R@1 & AP & R@1 & AP & R@1 & AP\\ \hline
  
  Backbone&Supervised& 28.59      &87.80  &87.80  &94.00  & 95.22  &96.52  &97.19 &96.32 &96.32 &96.25 &87.22 &97.50 &92.94 &98.75 &94.92 &97.50 &94.66 \\\hline
   \multirow{2}{*}{DMNIL} &Self-Supervised& 28.59  & 87.00  & 89.20   & 93.98   & 95.11  &96.50  & 97.23  &97.23   &97.86  & 95.00   & 85.46   & 97.50   & 92.72  &{100.00}  & 95.43  &{100.00}   &96.39\\
  &Supervised & 28.59   & \textbf{94.00}   & \textbf{95.21}   & \textbf{97.83}   & \textbf{98.88}  &\textbf{99.14}  & \textbf{95.44}  &\textbf{99.25}   &\textbf{99.39}    & \textbf{97.50}   & \textbf{92.71}   & \textbf{97.50}   &  \textbf{97.43}  &\textbf{100.00}  & \textbf{98.98}  &\textbf{98.75}   &\textbf{98.25} \\\hline \hline
\end{tabular}}
\end{table*}
\begin{figure*}[ht]
  \includegraphics[width=7.0in]{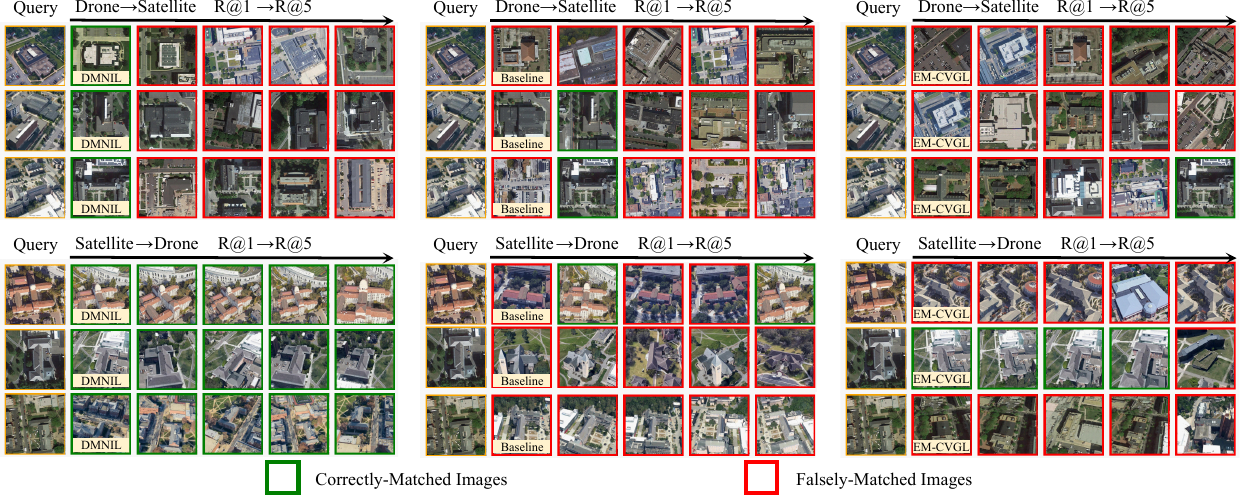}
  \caption{\textbf{Retrieval Results}. Retrieval examples of DMNIL, Baseline, and EM-CVGL on the University-1652 dataset. R@1$\rightarrow$R@5 results are shown for satellite$\rightarrow$drone and drone$\rightarrow$satellite scenarios. Green boxes represent correct matches and red boxes represent errors.}
  \label{fig5}
  \end{figure*}

\begin{figure}[ht]
\centering
  \includegraphics[width=3.4in]{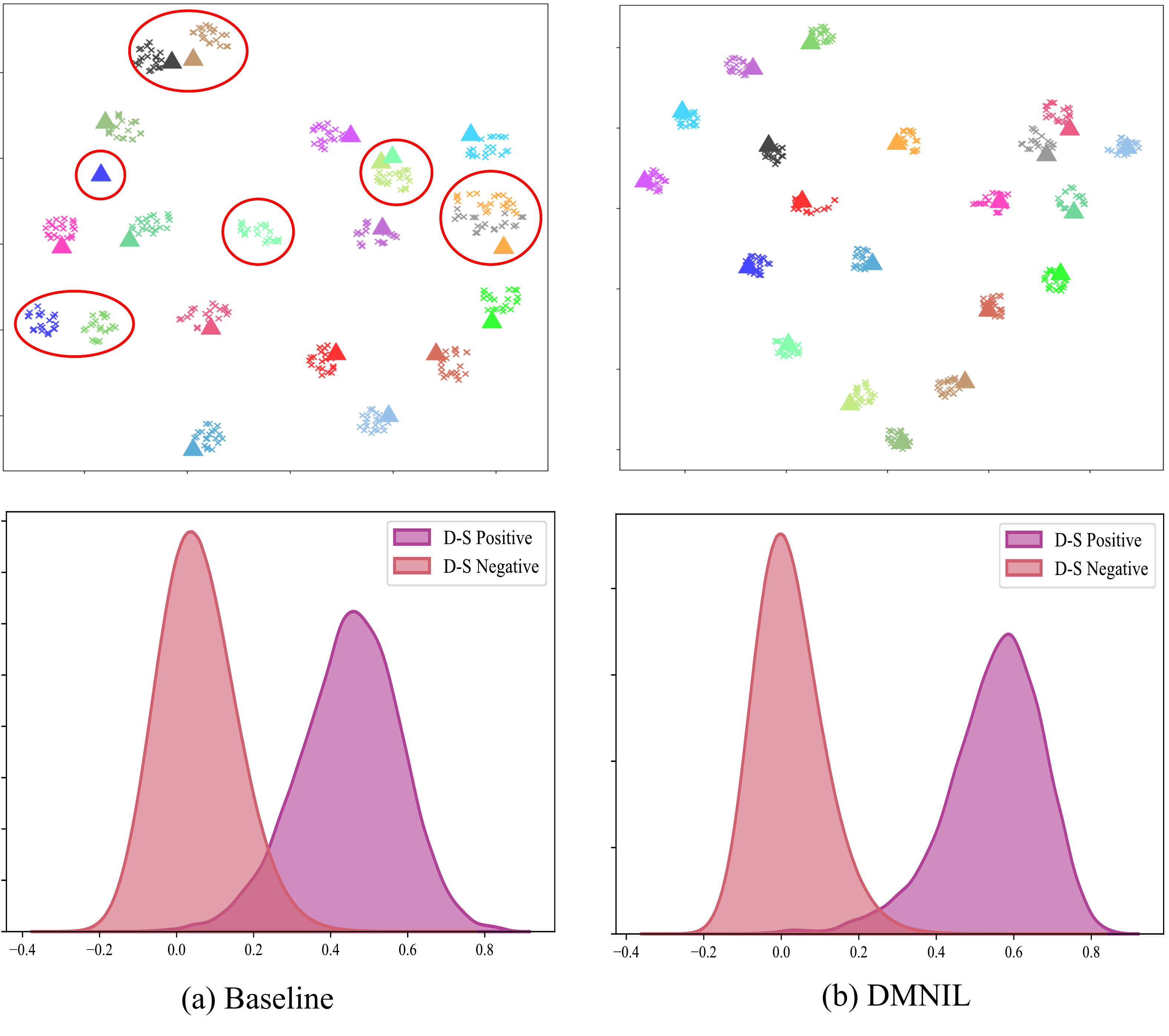}
  \caption{\textbf{Feature Space and Similarity Distribution}. The first row presents the t-SNE\cite{van2008visualizing} visualization, while the second row illustrates the similarity distribution for randomly selected geographic locations. In the t-SNE visualization, each color represents a distinct location, where '×' markers correspond to the drone view, and the triangles indicate the satellite-view of the same location.}
  \label{fig3}
  \end{figure}
{\color{black}\textit{Analysis of Unsupervised Learning}. To further explore the advantages of self-supervised learning, three configurations are reported for comparisons. In supervised learning type, the backbone of DMNIL is trained solely with an InfoNCE loss using paired data. In self-supervised learning type, the DMNIL employs the same backbone that is trained by $\mathcal{L}_{total}$ in Eq.~(\ref{eq23}) without any paired labels, which is namely the proposed method in this work. Furthermore, in another supervised learning type that inspired by [38], the DMNIL employs the same backbone trained with InfoNCE and $\mathcal{L}_{total}$ using both paired data and unpaired data.  As shown in Tables \ref{tab:learning configurations1} and \ref{tab:learning configurations2}, the supervised learning type of backbone achieves performance comparable to the proposed method in this work, which benefits from the paired relationship that is unavailable in self-supervised scenarios. It is crucial that the proposed method attains this accuracy without requiring pair data, which has demonstrated that its capacity to effectively exploit latent information from unpaired data. It is a significant advance for real-world applications where paired data is severely scarce.

The supervised type of DMNIL further enhances the performance and  obtains the latent information outside of the paired relationship. This demonstrates that the information learned from unpaired data is not equivalent to the paired data, and their combination gives additional performance improvement. Nevertheless, the fact that self-supervision can extract latent information from unpaired data highlights its unique value, in which such information cannot be fully replaced by explicit pairing cues. This phenomenon is highly interesting, and we will further explore the mechanisms behind it in future works.}

\subsection{Visualization}
\textit{Feature space and similarity distribution}. To further illustrate the effectiveness of the method, we perform a feature space and similarity distribution visualization for DMNIL, as presented in Fig. \ref{fig3}. Comparing (a) Baseline and (b) DMNIL, it is evident that DMNIL enhances the clustering of cross-view positive samples, effectively bringing the drone and satellite representations of the same location closer together. Additionally, the separation between different locations is more distinct, which indicates better feature discriminability. The second row shows the similarity distribution between drone-satellite positive and negative pairs. DMNIL demonstrates a clearer distinction between positive and negative pairs, with a reduced overlap between distributions, which suggests improved cross-view matching and reduced feature ambiguity. This highlights the effectiveness of DMNIL.

\begin{figure}[ht]
  \centering
  \includegraphics[width=3.4in]{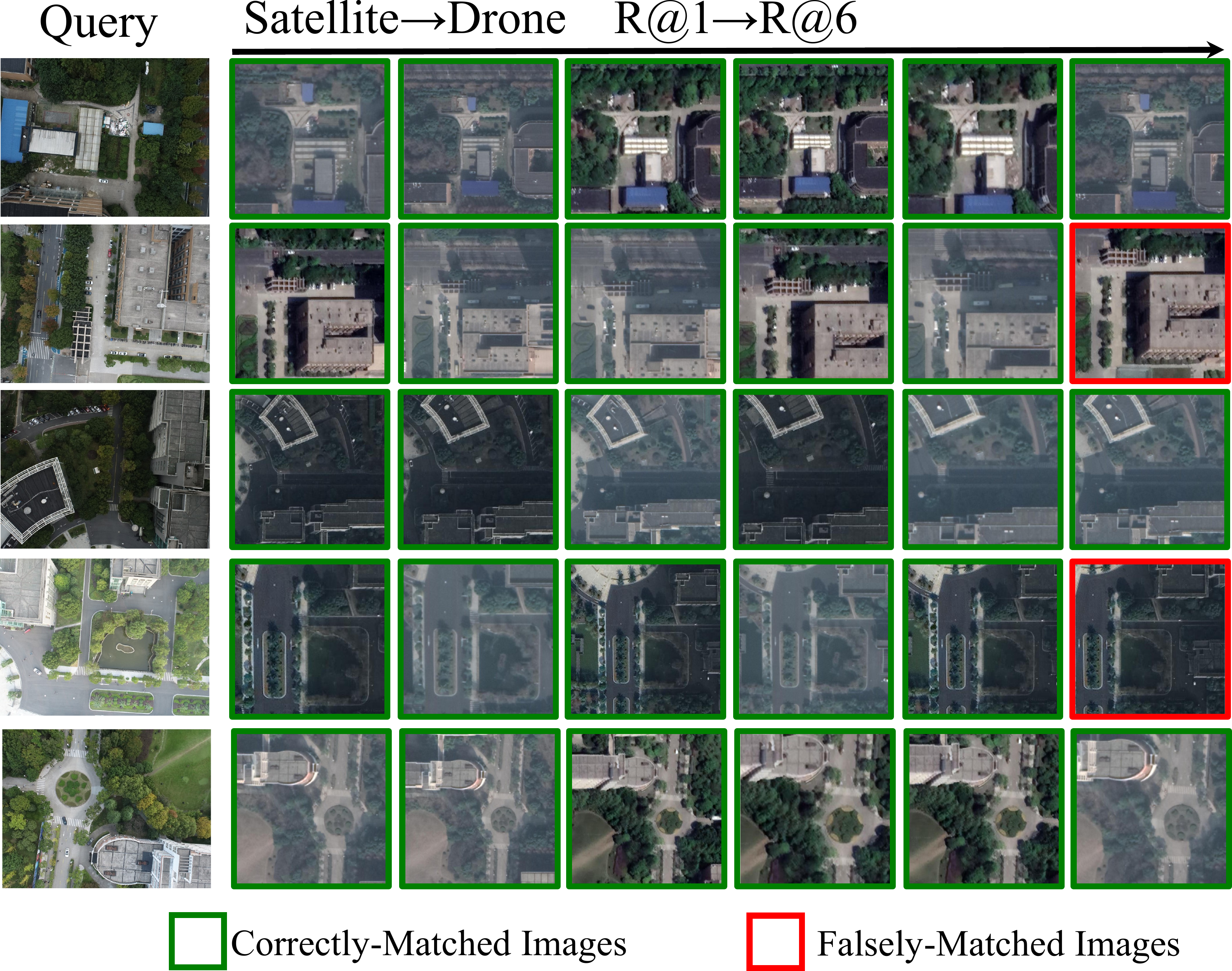}
  \caption{\textbf{Retrieval Results on DenseUAV}. R@1$\rightarrow$R@6 retrieval results of the proposed method on the DenseUAV dataset. The green box indicates a correctly-matched image, and the red box indicates a falsely-matched image.}
  \label{fig6}
  \end{figure}

\textit{Retrieval results}. {\color{black} To demonstrate the effectiveness of DMNIL, we present retrieval results on the University-1652 and DenseUAV datasets in Fig.~\ref{fig5} and Fig.~\ref{fig6}. On University-1652, for the drone$\rightarrow$satellite where each drone image has a single ground-truth satellite match, DMNIL retrieves the correct target at R@1 across all three locations, while both the Baseline and EM-CVGL fail to do so. For the satellite$\rightarrow$drone task, DMNIL also retrieves the correct drone-view images accurately, showing strong cross-view retrieval ability, whereas Baseline and EM-CVGL perform noticeably worse. Compared with University-1652, the DenseUAV dataset poses more realistic challenges with dense spatial sampling, multi-scale resolutions, and temporal dynamics. Despite the increased ambiguity, DMNIL effectively captures spatial geographic cues and achieves high retrieval accuracy, demonstrating robust cross-view performance under complex real-world conditions.}

\section{CONCLUSION}\label{conclusions}
This paper investigates a highly valuable and challenging DVGL task to reduce reliance on paired drone–satellite images. To this end, we propose a self-supervised DMNIL method that employs a shallow backbone with only 9 layers for feature extraction. Pseudo-labels are generated via clustering, and a dual-branch contrastive learning mechanism is introduced to enhance the capability in representing intra-view visual patterns. We design a DHML module to further improve intra-view discriminability and consistency. Additionally, to mitigate the structural and semantic domain gap between views, we propose an ICEL module, which facilitates semantic alignment by learning high-dimensional feature consistency across-views. To improve the quality and robustness of pseudo-labels, we introduce a pseudo-label enhancement strategy, which refines pseudo-supervision signals and boosts cross-view representation learning. As a result, the proposed DMNIL method effectively unifies both intra-view and cross-view feature representations in the embedding space. Extensive experiments conducted on three public benchmarks, including University-1652, SUES-200, and DenseUAV, demonstrate that DMNIL outperforms existing self-supervised methods, and even surpasses several fully supervised methods in certain scenarios without relying on paired drone–satellite images. These results validate the effectiveness and generalization capability of DMNIL and verify the potential for real-world DVGL applications.

\bibliographystyle{IEEEtran}
\footnotesize
\bibliography{reference}

@String(AAAI = {AAAI})

@inproceedings{deuser2023sample4geo,
  title={Sample4{G}eo: {H}ard negative sampling for cross-view geo-localisation},
  author={Deuser, Fabian and Habel, Konrad and Oswald, Norbert},
  booktitle = {Proc. IEEE Int. Conf. Comput. Vis.},
  year= {2023},
  pages= {16847-16856}
}

@article{shen2023mccg,
  title={{MCCG}: {A} convNeXt-based multiple-classifier method for cross-view geo-localization},
  author={Shen, Tianrui and Wei, Yingmei and Kang, Lai and Wan, Shanshan and Yang, Yee-Hong},
  journal={{IEEE} Trans. Circuits Syst. Video Technol.},
  volume={34},
  number={3},
  pages={1456-1468},
  year={2023},
  publisher={IEEE}
}

@article{chen2024sdpl,
  title={{SDPL}: {S}hifting-dense partition learning for {UAV}-view geo-localization},
  author={Chen, Quan and Wang, Tingyu and Yang, Zihao and Li, Haoran and Lu, Rongfeng and Sun, Yaoqi and Zheng, Bolun and Yan, Chenggang},
  journal={{IEEE} Trans. Circuits Syst. Video Technol.}, 
  volume={34},
  number={11},
  pages={11810-11824},
  year={2024},
  publisher={IEEE}
}

@article{ge2024multibranch,
  title={Multibranch joint representation learning based on information fusion strategy for cross-view geo-localization},
  author={Ge, Fawei and Zhang, Yunzhou and Liu, Yixiu and Wang, Guiyuan and Coleman, Sonya and Kerr, Dermot and Wang, Li},
  journal={{IEEE} Trans. Geosci. Remote Sens.},
  volume={62},
  pages={1-16},
  year={2024},
  publisher={IEEE}
}

@article{wang2021each,
  title={Each part matters: {L}ocal patterns facilitate cross-view geo-localization},
  author={Wang, Tingyu and Zheng, Zhedong and Yan, Chenggang and Zhang, Jiyong and Sun, Yaoqi and Zheng, Bolun and Yang, Yi},
  journal={{IEEE} Trans. Circuits Syst. Video Technol.},
  volume={32},
  number={2},
  pages={867-879},
  year={2021},
  publisher={IEEE}
}

@inproceedings{zhu2022transgeo,
  title={{T}rans{G}eo: {T}ransformer is all you need for cross-view image geo-localization},
  author={Zhu, Sijie and Shah, Mubarak and Chen, Chen},           
  booktitle = { Proc. IEEE Conf. Comput. Vis. Pattern Recognit.},
  year      = {2022},
  pages     = {1162-1171}
}

@article{xia2024enhancing,
  title={Enhancing cross-view geo-localization with domain alignment and scene consistency},
  author={Xia, Panwang and Wan, Yi and Zheng, Zhi and Zhang, Yongjun and Deng, Jiwei},
  journal={{IEEE} Trans. Circuits Syst. Video Technol.},
  year={2024},
  volume={34},
  number={12},
  pages={1-12},
  publisher={IEEE}
}

@inproceedings{liu2022convnet,
  title={A convNet for the 2020s},
  author={Liu, Zhuang and Mao, Hanzi and Wu, Chao-Yuan and Feichtenhofer, Christoph and Darrell, Trevor and Xie, Saining},
  booktitle = {Proc. IEEE Conf. Comput. Vis. Pattern Recognit.},
  year= {2022},
  pages = {11976-11986}

}

@article{du2024ccr,
  title={{CCR}: A counterfactual causal reasoning-based method for cross-view geo-localization},
  author={Du, Haolin and He, Jingfei and Zhao, Yuanqing},
  journal={{IEEE} Trans. Circuits Syst. Video Technol.},
  year={2024},
  volume={34},
  number={11},
  pages={11630-11643},
}

@article{wu2024camp,
  title={{CAMP}: {A}cross-view geo-localization method using contrastive attributes mining and position-aware partitioning},
  author={Wu, Qiong and Wan, Yi and Zheng, Zhi and Zhang, Yongjun and Wang, Guangshuai and Zhao, Zhenyang},
  journal={{IEEE} Trans. Geosci. Remote Sens.},
  year={2024},
  volume={62},
  number={},
  pages={1-14},
}

@article{lv2024direction,
  title={Direction-guided multi-scale feature fusion network for geo-localization},
  author={Lv, Hongxiang and Zhu, Hai and Zhu, Runzhe and Wu, Fei and Wang, Chunyuan and Cai, Meiyu and Zhang, Kaiyu},
  journal={{IEEE} Trans. Geosci. Remote Sens.},
  year={2024},
  volume={62},
  number={},
  pages={1-13},
  publisher={IEEE}
}

@article{wang2024multiple,
  title={Multiple-environment self-adaptive network for aerial-view geo-localization},
  author={Wang, Tingyu and Zheng, Zhedong and Sun, Yaoqi and Yan, Chenggang and Yang, Yi and Chua, Tat-Seng},
  journal={Pattern Recognit.},
  volume={152},
  pages={110363},
  year={2024},
  publisher={Elsevier}
}

@article{zhu2023sues,
  title={{SUES}-200: {A} multi-height multi-scene cross-view image benchmark across drone and satellite},
  author={Zhu, Runzhe and Yin, Ling and Yang, Mingze and Wu, Fei and Yang, Yuncheng and Hu, Wenbo},
  journal={{IEEE} Trans. Circuits Syst. Video Technol.},
  volume={33},
  number={9},
  pages={4825-4839},
  year={2023},
  publisher={IEEE}
}

@inproceedings{zheng2020university,
author = {Zheng, Zhedong and Wei, Yunchao and Yang, Yi},
title = {University-1652: {A} Multi-view Multi-source Benchmark for Drone-based Geo-localization},
year = {2020},
booktitle = {Proc. ACM Int. Conf. Multimedia},
pages = {1395–1403},
}

@inproceedings{liu2022swin,
  title={Swin transformer v2: {S}caling up capacity and resolution},
  author={Liu, Ze and Hu, Han and Lin, Yutong and Yao, Zhuliang and Xie, Zhenda and Wei, Yixuan and Ning, Jia and Cao, Yue and Zhang, Zheng and Dong, Li and others},
    booktitle = {Proc. IEEE Conf. Comput. Vis. Pattern Recognit.},
    year      = {2022},
    pages     = {12009-12019}
}

@article{van2008visualizing,
  title={{V}isualizing data using {t-SNE}.},
  author={Van der Maaten, Laurens and Hinton, Geoffrey},
  journal={J. Mach. Learn. Res.},
  volume={9},
  number={11},
  pages={2579-2605},
  year={2008}
}

@INPROCEEDINGS{2019Lending,
  author={Liu, Liu and Li, Hongdong},
  booktitle={Proc. IEEE Conf. Comput. Vis. Pattern Recognit.}, 
  title={Lending orientation to neural networks for cross-view geo-localization}, 
  year={2019},
  volume={},
  number={},
  pages={5617-5626}}

@ARTICLE{chen2024multi,
  author={Chen, Zhongwei and Yang, Zhao-Xu and Rong, Hai-Jun},
  journal={{IEEE} Trans. Geosci. Remote Sens.}, 
  title={Multilevel Embedding and Alignment Network With Consistency and Invariance Learning for Cross-View Geo-Localization}, 
  year={2025},
  volume={63},
  number={},
  pages={1-15}
 }

@article{DenseUAV,
  author={Dai, Ming and Zheng, Enhui and Feng, Zhenhua and Qi, Lei and Zhuang, Jiedong and Yang, Wankou},
  journal={{IEEE} Trans. Image Process.},
  title={Vision-Based UAV Self-Positioning in Low-Altitude Urban Environments},
  year={2024},
  volume={33},
  number={},
  pages={493-508}
}

@article{yin2023real,
  title={A real-time memory updating strategy for unsupervised person re-identification},
  author={Yin, Junhui and Zhang, Xinyu and Ma, Zhanyu and Guo, Jun and Liu, Yifan},
  journal={{IEEE} Trans. Image Process.},
  volume={32},
  pages={2309-2321},
  year={2023},
  publisher={IEEE}
}

@inproceedings{dai2022cluster,
  title={Cluster contrast for unsupervised person re-identification},
  author={Dai, Zuozhuo and Wang, Guangyuan and Yuan, Weihao and Zhu, Siyu and Tan, Ping},
  booktitle={Proc. Asian Conf. Comput. Vis.},
  pages={1142-1160},
  year={2022}
}

@inproceedings{ester1996density,
  title={A density-based algorithm for discovering clusters in large spatial databases with noise},
  author={Ester, Martin and Kriegel, Hans-Peter and Sander, J{\"o}rg and Xu, Xiaowei and others},
  booktitle={Proc. ACM SIGKDD Int. Conf. Knowl. Discov. Data Min.},
  volume={96},
  number={34},
  pages={226-231},
  year={1996}
}

@inproceedings{macqueen1967some,
  title={Some methods for classification and analysis of multivariate observations},
  author={MacQueen, James and others},
  booktitle={Proc. Fifth Berkeley Symp. Math. Statist. Prob.},
  volume={1},
  number={14},
  pages={281-297},
  year={1967},
}

@InProceedings{Li_2024_CVPR,
    author    = {Li, Guopeng and Qian, Ming and Xia, Gui-Song},
    title     = {Unleashing Unlabeled Data: A Paradigm for Cross-View Geo-Localization},
    booktitle = {Proc. IEEE Conf. Comput. Vis. Pattern Recognit.},
    year      = {2024},
    pages     = {16719-16729}
}

@inproceedings{adca,
  title={Augmented Dual-Contrastive Aggregation Learning for Unsupervised Visible-Infrared Person Re-Identification},
  author={Yang, Bin and Ye, Mang and Chen, Jun and Wu, Zesen},
  pages = {2843–2851},
  booktitle = {Proc. ACM Int. Conf. Multimedia},
  year={2022}
}

@InProceedings{Hu_2018_CVPR,
author = {Hu, Sixing and Feng, Mengdan and Nguyen, Rang M. H. and Lee, Gim Hee},
title = {CVM-Net: Cross-View Matching Network for Image-Based Ground-to-Aerial Geo-Localization},
booktitle = {Proc. IEEE Conf. Comput. Vis. Pattern Recognit.},
pages = {7258-7267},
year = {2018}
}

@inproceedings{zhang2023cross,
  title={{C}ross-view geo-localization via learning disentangled geometric layout correspondence},
  author={Zhang, Xiaohan and Li, Xingyu and Sultani, Waqas and Zhou, Yi and Wshah, Safwan},
  booktitle={Proc. AAAI Conf. Artif. Intell.},
  volume={37},
  number={3},
  pages={3480-3488},
  year={2023}
}

@InProceedings{Zhu_2017_ICCV,
author = {Zhu, Jun-Yan and Park, Taesung and Isola, Phillip and Efros, Alexei A.},
title = {Unpaired Image-To-Image Translation Using Cycle-Consistent Adversarial Networks},
booktitle = {Proc. IEEE Int. Conf. Comput. Vis.},
pages={2223--2232},
year = {2017}
}

@inproceedings{yang2024shallow,
  title={Shallow-Deep Collaborative Learning for Unsupervised Visible-Infrared Person Re-Identification},
  author={Yang, Bin and Chen, Jun and Ye, Mang},
  booktitle={Proc. IEEE Conf. Comput. Vis. Pattern Recognit.},
  pages={16870-16879},
  year={2024}
}

@InProceedings{He_2020_CVPR,
author = {He, Kaiming and Fan, Haoqi and Wu, Yuxin and Xie, Saining and Girshick, Ross},
title = {Momentum Contrast for Unsupervised Visual Representation Learning},
booktitle = {Proc. IEEE Conf. Comput. Vis. Pattern Recognit.},
year = {2020},
pages = {9729-9738}
}

@InProceedings{Wu_2018_CVPR,
author = {Wu, Zhirong and Xiong, Yuanjun and Yu, Stella X. and Lin, Dahua},
title = {Unsupervised Feature Learning via Non-Parametric Instance Discrimination},
booktitle = {Proc. IEEE Conf. Comput. Vis. Pattern Recognit.},
pages={3733-3742},
year = {2018}
}

@InProceedings{Yang_2023_ICCV,
    author    = {Yang, Bin and Chen, Jun and Ye, Mang},
    title     = {Towards Grand Unified Representation Learning for Unsupervised Visible-Infrared Person Re-Identification},
    booktitle = {Proc. IEEE Int. Conf. Comput. Vis.},
    year      = {2023},
    pages     = {11069-11079}
}

@article{yang2023dual,
  title={Dual consistency-constrained learning for unsupervised visible-infrared person re-identification},
  author={Yang, Bin and Chen, Jun and Chen, Cuiqun and Ye, Mang},
  journal={IEEE Trans. Inf. Forensics Secur.},
  volume={19},
  pages={1767--1779},
  year={2023},
  publisher={IEEE}
}

@inproceedings{hu2022feature,
  title={Feature Representation Learning for Unsupervised Cross-domain Image Retrieval},
  author={Hu, Conghui and Lee, Gim Hee},
  booktitle={Proc. Eur. Conf. Comput. Vis.},
  year={2022},
  pages = {529–544}
}

@inproceedings{deng2009imagenet,
  title={Imagenet: A large-scale hierarchical image database},
  author={Deng, Jia and Dong, Wei and Socher, Richard and Li, Li-Jia and Li, Kai and Fei-Fei, Li},
  booktitle={Proc. IEEE Conf. Comput. Vis. Pattern Recognit.},
  pages={248-255},
  year={2009}
}

@inproceedings{wei2018person,
  title={Person transfer gan to bridge domain gap for person re-identification},
  author={Wei, Longhui and Zhang, Shiliang and Gao, Wen and Tian, Qi},
  booktitle={Proc. IEEE Conf. Comput. Vis. Pattern Recognit.},
  pages={79--88},
  year={2018}
}

@article{zheng2017discriminatively,
  title={A discriminatively learned cnn embedding for person reidentification},
  author={Zheng, Zhedong and Zheng, Liang and Yang, Yi},
  journal={ACM Trans. Multimedia Comput. Commun. Appl.},
  volume={14},
  number={1},
  pages={1-20},
  year={2017},
}

@article{sun2014deep,
  title={Deep learning face representation by joint identification-verification},
  author={Sun, Yi and Chen, Yuheng and Wang, Xiaogang and Tang, Xiaoou},
  journal={Proc. Adv. Neural Inf. Process. Syst.},
  volume={27},
  pages={1988-1996},
  year={2014}
}

@inproceedings{zheng2015scalable,
  title={Scalable person re-identification: A benchmark},
  author={Zheng, Liang and Shen, Liyue and Tian, Lu and Wang, Shengjin and Wang, Jingdong and Tian, Qi},
  booktitle={Proc. IEEE Int. Conf. Comput. Vis.},
  pages={1116-1124},
  year={2015}
}

@inproceedings{shi2019spatial,
  title={Spatial-aware feature aggregation for image based cross-view geo-localization},
  author={Shi, Yujiao and Liu, Liu and Yu, Xin and Li, Hongdong},
  booktitle={Proc. Adv. Neural Inf. Process. Syst.},
  pages= {10090–10100},
  year={2019}
}

@ARTICLE{10601183,
  author={Ahn, Woo-Jin and Park, So-Yeon and Pae, Dong-Sung and Choi, Hyun-Duck and Lim, Myo-Taeg},
  journal={{IEEE} Trans. Geosci. Remote Sens.}, 
  title={Bridging Viewpoints in Cross-View Geo-Localization With Siamese Vision Transformer}, 
  year={2024},
  volume={62},
  number={},
  pages={1-12},
  }

@inproceedings{arandjelovic2016netvlad,
  title={NetVLAD: CNN architecture for weakly supervised place recognition},
  author={Arandjelovic, Relja and Gronat, Petr and Torii, Akihiko and Pajdla, Tomas and Sivic, Josef},
  booktitle={Proc. IEEE Conf. Comput. Vis. Pattern Recognit.},
  pages={5297--5307},
  year={2016}
}

@inproceedings{zhai2017predicting,
  title={Predicting ground-level scene layout from aerial imagery},
  author={Zhai, Menghua and Bessinger, Zachary and Workman, Scott and Jacobs, Nathan},
  booktitle={Proc. IEEE Conf. Comput. Vis. Pattern Recognit.},
  pages={867--875},
  year={2017}
}

@inproceedings{zhu2021vigor,
  title={Vigor: Cross-view image geo-localization beyond one-to-one retrieval},
  author={Zhu, Sijie and Yang, Taojiannan and Chen, Chen},
  booktitle={Proc. IEEE Conf. Comput. Vis. Pattern Recognit.},
  pages={3640--3649},
  year={2021}
}

@ARTICLE{li2024learning,
  author={Li, Haoyuan and Xu, Chang and Yang, Wen and Yu, Huai and Xia, Gui-Song},
  journal={{IEEE} Trans. Geosci. Remote Sens.}, 
  title={Learning Cross-View Visual Geo-Localization Without Ground Truth}, 
  year={2024},
  volume={62},
  number={},
  pages={1-17}
}

@article{oquab2023dinov2,
  title={Dinov2: Learning robust visual features without supervision},
  author={Oquab, Maxime and Darcet, Timoth{\'e}e and Moutakanni, Th{\'e}o and Vo, Huy and Szafraniec, Marc and Khalidov, Vasil and Fernandez, Pierre and Haziza, Daniel and Massa, Francisco and El-Nouby, Alaaeldin and others},
  journal={arXiv preprint arXiv:2304.07193},
  year={2023}
}

@article{chen2025limited,
  title={From Limited Labels to Open Domains: An Efficient Learning Paradigm for UAV-view Geo-Localization},
  author={Chen, Zhongwei and Yang, Zhao-Xu and Rong, Hai-Jun and Lang, Jiawei},
  journal={arXiv preprint arXiv:2503.07520},
  year={2025}
}

@inproceedings{chen2021ice,
  title={Ice: Inter-instance contrastive encoding for unsupervised person re-identification},
  author={Chen, Hao and Lagadec, Benoit and Bremond, Francois},
  booktitle={Proc. IEEE Int. Conf. Comput. Vis.},
  pages={14960--14969},
  year={2021}
}

@inproceedings{wang2022optimal,
  title={Optimal transport for label-efficient visible-infrared person re-identification},
  author={Wang, Jiangming and Zhang, Zhizhong and Chen, Mingang and Zhang, Yi and Wang, Cong and Sheng, Bin and Qu, Yanyun and Xie, Yuan},
  booktitle={Proc. Eur. Conf. Comput. Vis.},
  pages={93--109},
  year={2022}
}

@ARTICLE{10882953,
  author={Ye, Mang and Wu, Zesen and Du, Bo},
  journal={{IEEE} Trans. Pattern Anal. Mach. Intell.}, 
  title={Dual-Level Matching with Outlier Filtering for Unsupervised Visible-Infrared Person Re-Identification}, 
  year={2025},
  volume={},
  number={},
  pages={1-14}}

@inproceedings{cheng2023efficient,
  title={Efficient bilateral cross-modality cluster matching for unsupervised visible-infrared person reid},
  author={Cheng, De and He, Lingfeng and Wang, Nannan and Zhang, Shizhou and Wang, Zhen and Gao, Xinbo},
  booktitle={Proc. ACM Int. Conf. Multimedia},
  pages={1325--1333},
  year={2023}
}

@ARTICLE{9840394,
  author={Zhang, Hongwei and Zhang, Guoqing and Chen, Yuhao and Zheng, Yuhui},
  journal={{IEEE} Trans. Circuits Syst. Video Technol.}, 
  title={Global Relation-Aware Contrast Learning for Unsupervised Person Re-Identification}, 
  year={2022},
  volume={32},
  number={12},
  pages={8599-8610}}

@inproceedings{wang2025coarse,
  title={From Coarse to Fine: A Matching and Alignment Framework for Unsupervised Cross-View Geo-Localization},
  author={Wang, Xueyi and Zhang, Lele and Fan, Zheng and Liu, Yang and Chen, Chen and Deng, Fang},
  booktitle={Proc. AAAI Conf. Artif. Intell.},
  volume={39},
  number={8},
  pages={8024--8032},
  year={2025}
}

@ARTICLE{10680455,
  author={Ji, Haoxuanye and Wang, Le and Zhou, Sanping and Tang, Wei and Zheng, Nanning and Hua, Gang},
  journal={IEEE Trans. Neural Networks Learn. Syst.}, 
  title={Meta Pairwise Relationship Distillation for Unsupervised Person Re-Identification}, 
  year={2025},
  volume={36},
  number={5},
  pages={9165-9179}}

@ARTICLE{9457243,
  author={Zhang, Liyan and Du, Guodong and Liu, Fan and Tu, Huawei and Shu, Xiangbo},
  journal={IEEE Trans. Neural Networks Learn. Syst.}, 
  title={Global–Local Multiple Granularity Learning for Cross-Modality Visible–Infrared Person Reidentification}, 
  year={2025},
  volume={36},
  number={3},
  pages={4209-4219}}

@ARTICLE{10310270,
  author={Wu, Qiong and Li, Jiahan and Dai, Pingyang and Ye, Qixiang and Cao, Liujuan and Wu, Yongjian and Ji, Rongrong},
  journal={IEEE Trans. Neural Networks Learn. Syst.}, 
  title={Unsupervised Domain Adaptation on Person Reidentification Via Dual-Level Asymmetric Mutual Learning}, 
  year={2025},
  volume={36},
  number={1},
  pages={1371-1382}}

@article{mcinnes2017hdbscan,
  title={hdbscan: Hierarchical density based clustering.},
  author={McInnes, Leland and Healy, John and Astels, Steve and others},
  journal={J. Open Source Softw.},
  volume={2},
  number={11},
  pages={205},
  year={2017}
}

@article{ward1963hierarchical,
  title={Hierarchical grouping to optimize an objective function},
  author={Ward Jr, Joe H},
  journal={J. Am. Stat. Assoc.},
  volume={58},
  number={301},
  pages={236--244},
  year={1963},
  publisher={Taylor \& Francis}
}

\end{document}